\documentclass[preprint,12pt]{elsarticle}

\usepackage{amssymb}
\usepackage{amsfonts,amsmath,amscd}
\usepackage{float}
\usepackage{graphicx}
\usepackage{subfig}
\usepackage{nccmath}
\usepackage{siunitx}
\usepackage{booktabs}
\usepackage{url}
\usepackage{caption}
\usepackage{algorithm}
\usepackage{algpseudocode}

\journal{Knowledge-Based Systems}

\begin{document}

\begin{frontmatter}

\title{Improving Safety in Mixed Traffic: A Learning-based Model Predictive Control for Autonomous and Human-Driven Vehicle Platooning}

\author[label1]{Jie\ Wang\corref{cor1}}
\ead{jwangjie@outlook.com}
\cortext[cor1]{Corresponding author}
\author[label2]{Zhihao\ Jiang}
\ead{jiangzhh@shanghaitech.edu.cn}
\author[label1]{Yash Vardhan\ Pant}
\ead{yash.pant@uwaterloo.ca}

\affiliation[label1]{organization={Electrical and Computer Engineering Department, University of Waterloo},
            addressline={200 University Avenue West}, 
            city={Waterloo},
            postcode={N2L~3G1}, 
            state={ON},
            country={Canada}}

\affiliation[label2]{organization={School of Information Science and Technologies, ShanghaiTech University},
            addressline={393 Middle Huaxia Road}, 
            city={Pudong},
            postcode={201210}, 
            state={Shanghai},
            country={China}}

\begin{abstract}
As autonomous vehicles (AVs) become more common on public roads, their interaction with human-driven vehicles (HVs) in mixed traffic is inevitable. This requires new control strategies for AVs to handle the unpredictable nature of HVs. This study focused on safe control in mixed-vehicle platoons consisting of both AVs and HVs, particularly during longitudinal car-following scenarios. We introduce a novel model that combines a conventional first-principles model with a Gaussian process (GP) machine learning-based model to better predict HV behavior. Our results showed a significant improvement in predicting HV speed, with a 35.64\% reduction in the root mean square error compared with the use of the first-principles model alone. We developed a new control strategy called GP-MPC, which uses the proposed HV model for safer distance management between vehicles in the mixed platoon. The GP-MPC strategy effectively utilizes the capacity of the GP model to assess uncertainties, thereby significantly enhancing safety in challenging traffic scenarios, such as emergency braking scenarios. In simulations, the GP-MPC strategy outperformed the baseline MPC method, offering better safety and more efficient vehicle movement in mixed traffic.
\end{abstract}


\end{frontmatter}

\section{Introduction}
\label{section:Introduction}

\begin{figure}
    \centerline{\includegraphics[width=0.98\columnwidth]{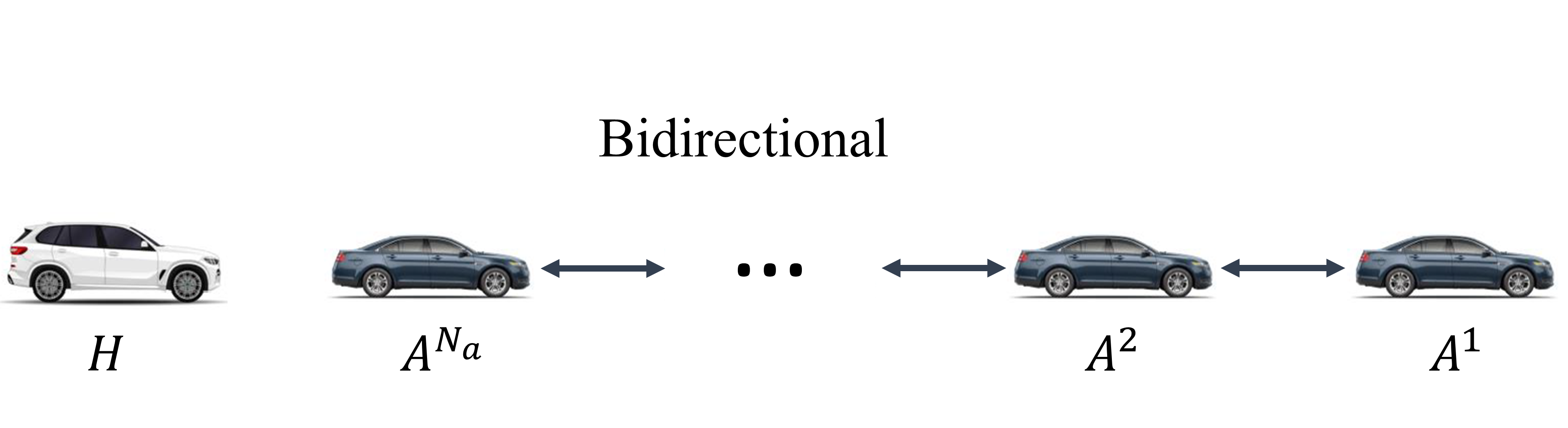}}
    \caption{Platoon of $N_a$ autonomous vehicles (AVs) depicted as $\{A^1, A^2, \cdots, A^{N_a}\}$ and a following human-driven vehicle (HV) $H$. The AVs in the platoon can exchange information through a bidirectional communication topology. This cooperative behavior contrasts with a single AV--HV scenario in which the traffic ahead is not cooperative. In an AV platoon, AVs can synchronize their behaviors to create safer interaction between the last AV and the trailing HV. 
    }
    \label{figure:1}
\end{figure}

Over the past decade, autonomous vehicles (AVs) have revolutionized the concept of intelligent transportation, resulting in advancements such as AV platooning \cite{guanetti2018}. This technology enhances the traffic flow by optimizing closely coordinated vehicle movements \cite{martinez2021}. Such coordination is facilitated by advanced communication technologies, including vehicle-to-infrastructure and vehicle-to-vehicle communication \cite{aramrattana2021}, which enable seamless and efficient interactions among vehicles. However, human-driven vehicles (HVs) are expected to continue to dominate roads for a foreseeable period \cite{rahmati2019}, resulting in frequent interactions between AVs and HVs. Fig. \ref{figure:1} illustrates a mixed-traffic scenario in which an HV follows an AV platoon.

A recent study focusing on traffic accidents involving AVs showed that 64.2\% of accidents in mixed-traffic scenarios involved HVs rear-ending AVs, a significant increase compared with 28.3\% in conventional HV-only traffic \cite{petrovic2020}. This suggests that drivers are less familiar with the dynamic behavior of AV platoons, which results in a higher risk of accidents. Addressing this challenge requires novel AV platoon-control strategies that consider human driver behavior, rather than relying solely on drivers to adapt to AVs. Previous research on HV--AV interactions in car-following scenarios has proposed several HV models \cite{sadat2020}. However, a major limitation of these models is their inability to adapt to practical model-based control systems. Thus, interpretable and implementable HV--AV interaction models that are suitable for model-based control systems are essential. These models should aim to enhance safety and achieve other desired outcomes in mixed-traffic environments.

\subsection{Contributions}

This paper proposes an innovative learning-based model predictive control (MPC) approach for the longitudinal car-following control of mixed-vehicle platoons. The developed MPC policy explicitly considers uncertainties associated with HVs to enhance safety during AV--HV interactions in complex traffic scenarios such as emergency braking. The following are the key contributions of this study:
\begin{itemize}
    \item The development of a novel HV modeling approach. This approach integrates a first-principles model, capturing human behaviors within a velocity tracking context \cite{macadam2003}, with a learning-based Gaussian process (GP) model. The GP model, trained on data from a human-in-the-loop simulator, refines the predictive velocity of the first-principles model, reducing the modeling error by approximately 35.6

    \item The formulation of a chance-constrained MPC strategy for mixed-vehicle platoons. This strategy considers various factors, including acceleration, velocity, and the requirement for safe distancing. This safe distance includes a pre-defined deterministic safe distance and an adaptive safe distance between an HV and the leading AV, which incorporates a probabilistic constraint derived from variance estimations provided by the GP models.

    \item Simulation studies verified the effectiveness of the proposed GP-MPC method over a baseline nominal MPC. The GP-MPC strategy ensures larger minimum distances between vehicles and facilitates higher travel speeds, thereby enhancing the safety and efficiency of the longitudinal car-following control of mixed-vehicle platoons.
\end{itemize}

\subsection{Related work}
\label{sec:related_work}

In the field of AV control, particularly mixed-vehicle platooning, adaptive cruise control (ACC) and cooperative adaptive cruise control (CACC) have progressed significantly \cite{yu2022researches}. ACC methods can efficiently maintain a stable temporal gap in the leading vehicle under uniform traffic conditions, based on a constant-time headway policy. However, its dependency on deterministic inputs limits its effectiveness in mixed-traffic scenarios, in which unpredictability and variability are prominent. By expanding on ACC, CACC enhances platoon formation and responsiveness to traffic dynamics by incorporating vehicle-to-vehicle communication. Although CACC improves on ACC by enabling a tighter platoon formation and quicker response to traffic dynamics, it is still limited by the inadequate prediction of HV behavior, which is critical in mixed-traffic environments \cite{li2022cooperative}.

Recently, a more advanced technique, MPC, has been increasingly applied to AV platoon systems \cite{yu2021}. Unlike ACC and CACC, which are reactive and based on immediate sensor inputs, MPC uses a model of vehicle dynamics and the surrounding environment to predict future states and optimize control actions over a moving horizon \cite{wang2023learning}. This predictive ability is particularly beneficial in mixed-traffic scenarios, in which HVs introduce uncertainty. By considering a range of future outcomes, MPC can simultaneously handle multiple objectives and constraints, such as safety, efficiency, and ride comfort, which ensures robust control strategies in AV operations that adapt to the variability of human driving behaviors.

Car-following models are essential for the development and implementation of ACC, CACC, and MPC. Among the established models, the constant speed model (CSM) is a straightforward approach that assumes constant vehicle speed for motion prediction and allows for direct speed measurement by surrounding vehicles. This simplicity makes it a valuable benchmark in mixed traffic studies \cite{guo2021anticipative}. Furthermore, the Gazis--Herman--Rothery (GHR) model is one of the earliest and foundational approaches. It defines a vehicle's acceleration based on its speed, speed difference from the leading vehicle, and headway (distance to the leading vehicle), providing a simple but effective framework for basic automated following behaviors. The intelligent driver model (IDM) is a more contemporary model that adapts a vehicle's acceleration by considering factors such as current and desired speeds, minimum safe distance to the leading vehicle, and relative speed. The full velocity difference model (FVDM) incorporates a more comprehensive consideration of the velocity differences between the following and leading vehicles. This enhanced approach is beneficial for vehicle platooning when the behaviors of multiple vehicles in a platoon must be harmonized for optimal flow and safety. These models focus on capturing the dynamics of the following vehicle based on the actions of the leading vehicle and emphasize reaction times, safe following distances, and relative velocities \cite{li2023survey}. However, the reliance of these models on acceleration data can be a disadvantage, particularly as acceleration data tends to be noisier than velocity data. An alternative for human-driven vehicle modeling in mixed traffic employs a second-order transfer function with velocity-based input-output relationships, supplemented by a time delay to mimic human reaction times \cite{macadam2003}. This velocity-focused approach sidesteps the issues inherent in acceleration-based models, offering a more robust framework for analyzing the intricate dynamics of mixed traffic environments.

However, when modeling HV--AV interactions in mixed-traffic scenarios, these conventional models assume identical behaviors in HV--AV interactions using fixed reaction delays for human drivers. Although these models offer a moderate representation of human driving patterns, their performance is limited by parameter constraints, which hinder their capacity to capture the intricate and complex behaviors exhibited by human drivers \cite{guo2020}. To bridge this gap, the field of car-following control has shifted from traditional parametric models to data-driven modeling methods. Researchers are increasingly exploring approaches that leverage machine-learning techniques to develop more accurate and adaptable models for modeling car-following behavior. Techniques based on artificial neural networks, such as radial basis function networks \cite{panwai2007}, multilayer neural networks \cite{khodayari2012}, and recurrent neural networks \cite{morton2016}, have been proposed to account for the immediate reaction delay of human drivers. Gaussian mixture regression \cite{lefevre2014a} and hidden Markov models \cite{qu2017} are other methodologies that can manage the stochastic uncertainties of human drivers. Nonparametric methods have demonstrated superior predictive accuracy in capturing human driver behavior compared with conventional parametric models \cite{lefevre2014b}. 

Among these learning-based models, GP regression is prominent for its ability to model complex systems and provide uncertainty assessments along with predictions \cite{wang2023learning}, which are crucial for mixed-traffic scenarios involving stochastic human behaviors. In the realm of learning-based robotic control, a system model is depicted by combining a nominal (parametric) and GP model, which encapsulates the deviations between the actual system behaviors and the parametric model \cite{hewing2020}. Because human behavior encompasses both deterministic and stochastic components \cite{chen2010}, we propose a novel approach for modeling HVs by merging a traditional parametric model with a GP model, thereby combining the advantages of both modeling paradigms. 

In the application of MPC to mixed traffic, various innovative approaches have been developed, each targeting specific challenges within this complex domain. The approach in \cite{guo2021anticipative} utilizes the inverse MPC to enhance the anticipation of HV states, particularly in scenarios constrained by communication limitations. This method demonstrates advancements in the prediction of vehicle behavior in mixed traffic. However, it primarily relies on inverse modeling techniques that do not estimate HV uncertainty or consider it in the design of the MPC policy. Another innovative approach is the Tube MPC framework \cite{feng2021robust}, which utilizes probabilistic bounds to manage the uncertainties related to HV behavior. Although this method ensures robust control strategies, it requires predefined uncertainty bounds, in contrast to our GP-MPC, which can dynamically estimate HV uncertainties. A recent approach using the Koopman operator theory in \cite{zhan2022data} models mixed-vehicle platoons using a linear model in a high-dimensional space. This approach, achieved through a neural network framework, addresses platoon-control problems with centralized and distributed MPC algorithms, which require significant computational resources owing to their high-dimensional modeling approach.

Because addressing the uncertainty of HVs is a major challenge in mixed-traffic management \cite{feng2021robust}, we propose a novel GP-based HV model to quantify the HV uncertainty and develop an MPC framework to handle the uncertainty. Compared with state-of-the-art MPC methods for managing mixed traffic, our GP-MPC offers a direct and dynamic estimation of HV uncertainties. This is in contrast to approaches, such as those in \cite{guo2021anticipative}, that utilize inverse modeling techniques to improve only the accuracy of the HV behavior estimation, and \cite{feng2021robust}, where the probabilistic uncertainty bounds are predefined. GP-MPC offers a more nuanced approach to handling the unpredictable and diverse behaviors of HVs, which is crucial for the realistic modeling of mixed traffic but is not explicitly addressed in the aforementioned methods. To the best of our knowledge, this is the first time that the quantified uncertainty of HVs has been explicitly utilized as a constraint in MPC policy design for platoon control in a mixed traffic flow.

The remainder of this paper is organized as follows: Sec. \ref{sec:hv_modelling} outlines the modeling of HVs for the car-following control. Sec. \ref{sec:control} develops the proposed GP learning-based MPC methodology (GP-MPC). Sec. \ref{sec:simulation} provides the simulation case studies comparing GP-MPC with a baseline MPC. Finally, Sec. \ref{sec:conclusion} offers concluding thoughts.

\section{Human-driven Vehicle Model}
\label{sec:hv_modelling}

This section outlines the development of the HV model, which combines a first-principles delay-inclusive transfer function (nominal model) with a GP learning-based model. This novel approach significantly improves the accuracy of HV behavior modeling and assesses modeling variance, which is crucial for safe operation in mixed-vehicle platoons.

\subsection{First-principles nominal model}
\label{sec:nominal_modeling}

Human reaction time delay is a critical factor that affects driver performance \cite{pirani2022}. To capture this, reference \cite{macadam2003} proposed a transfer function with a time delay model that incorporates the complexities of human central nervous system latencies, neuromuscular system, and other relevant human and environmental factors to model the human response as follows:
\begin{equation} 
    G_{H}(s) \approx K \frac{1+T_{z} s}{1+2 \gamma T_{w} s+T_{w}^{2} s^{2}} e^{-T_{d} s}=\frac{\dot{P}_{H}(s)}{\dot{P}_{N}(s)}  \, , \tag{1} \label{eqn:tf}
\end{equation}
Here, $G_{H}(s)$ is the transfer function in the s-domain, where $K$ denotes the system gain, $T_{z}$ is the zero-time constant, $\gamma$ is the damping coefficient, and $T_{w}$ is the natural frequency of the system. The term $e^{-T_{d} s}$ introduces a time delay into the system, where $T_{d}$ is the delay in the response of the human driver. The variables $\dot{P}_{H}(s)$ and $\dot{P}_{N}(s)$ are the Laplace transformations of the velocities for the HV $v_{k}^{H}$ and front AV $v_{k}^{N_a}$, respectively, which are utilized to transform the time-domain velocity signals into the s-domain. This transformation facilitates the analysis and design of control systems by converting differential equations into algebraic equations, making them easier to manipulate and solve.

To address the complexities introduced by the time delay in $G_{H}(s)$, we apply a second-order Padé approximation \cite{brezinski1994pade}. This method is selected for its precise representation of time-delay systems and computational simplicity. The discrete-time HV model derived from this approximation provides a balance between detail and manageability. Following the Padé approximation, we discretize the transfer function \eqref{eqn:tf} as follows:
\begin{equation}
    G(z)= \frac{V_{k}^{H}(z)}{V_{k}^{N_a}(z)} = \frac{b_{1} z^{3}+b_{2} z^{2}+ b_{3} z +b_{4}}{z^{4}+c_{1} z^{3}+c_{2} z^{2}+c_{3} z+c_{4}}  \, . \tag{2} \label{eqn:tf_z}
\end{equation}
Rearranging \eqref{eqn:tf_z}, we formulate a discrete-time difference equation for the autoregressive with exogenous input (ARX) model \cite{sekizawa2007modeling} as
\begin{ceqn} 
    \begin{align}
        v^{H}_{k} &= -c_1 v^{H}_{k-1} - c_2 v^{H}_{k-2} - c_3 v^{H}_{k-3} - c_4 v^{H}_{k-4} \nonumber \\
        & \, \quad + b_1 v^{N_a}_{k-1} + b_2 v^{N_a}_{k-2} + b_3 v^{N_a}_{k-3} + b_4 v^{N_a}_{k-4} \, , \nonumber \\
        &= {f}\left(v^{H}_{k-1}, v^{H}_{k-2}, v^{H}_{k-3}, v^{H}_{k-4}, v^{N_a}_{k-1}, v^{N_a}_{k-2}, v^{N_a}_{k-3}, v^{N_a}_{k-4} \right) \, , \nonumber \\ 
        & = {f}\left(v_{k-1:k-4}^{H}, {v}_{k-1:k-4}^{N_a} \right) \, . \tag{3} \label{eqn:arx}
    \end{align}
\end{ceqn}
where $v^{H}_{k-i}$ and ${v}^{N_a}_{k-i}$ represent the velocities of an HV and the last AV, respectively. 

Although the transfer function model with a reaction delay provides moderate accuracy in representing human driving behaviors, as shown in \cite{pirani2022}, it does not offer insights into the modeling uncertainties. Recognizing the importance of accurately modeling HVs and quantifying these uncertainties for safe platoon control, we consider GP models \cite{wang2023intuitive}. The GP models are used to identify the discrepancies between the ARX model \eqref{eqn:arx} and actual HV behaviors. This identification not only enhances the accuracy of the HV model but also offers a quantifiable measure of modeling uncertainties. These uncertainties are then incorporated as safety constraints into our control mechanisms, significantly enhancing the reliability of the human-in-the-loop platooning system.

\subsection{Gaussian process regression}
\label{section:gpr}

GP regression is a widely used machine-learning technique for learning-based control \cite{brunke2021}. Given an input dataset $\mathbf{a} = [\mathbf{a}_1, \cdots, \mathbf{a}_m]^\mathsf{T} \in \mathbb{R}^{n_a \times m}$ consisting of $m$ data points and their corresponding measurements $\mathbf{d} = [\mathbf{d}_1, \cdots, \mathbf{d}_m]^\mathsf{T} \in \mathbb{R}^{n_d \times m}$, the relationship between them can be expressed as $\mathbf{d}_k = \mathbf{g}(\mathbf{a}_k) + \boldsymbol{\omega}_k$, where $\boldsymbol{\omega}_k$ represents independent and identically distributed Gaussian noise with $\boldsymbol{\omega}_k \sim \mathcal{N} (0, \boldsymbol{\Sigma^\omega})$. The covariance matrix $\boldsymbol{\Sigma^\omega}$ is a diagonal matrix with entries $[\sigma_1^2, \cdots, \sigma_{n_d}^2]$. The function $\mathbf{g}(\cdot)$ can be estimated using the observed input--output dataset: 
\begin{ceqn}
    \begin{align}
       \mathcal{D}_m = \{\mathbf{a} = [\mathbf{a}_1, \cdots, \mathbf{a}_m]^\mathsf{T},  \mathbf{d} = [\mathbf{d}_1, \cdots, \mathbf{d}_m]^\mathsf{T} \} \, . \tag{4} \label{eqn:dataset}
    \end{align}
\end{ceqn}

Assuming that each output dimension of $\mathbf{d}_k$ is independent, given input $\mathbf{a}_k$, we correspond a GP with zero mean and a prior kernel $k^M(\cdot, \cdot)$ to every dimension $M$ within set $\{1, \cdots, n_d\}$. For each dimension, the measurement data denoted as $[\mathbf{d}_k]_{M, \: \cdot}$ follows a normal distribution represented as $[\mathbf{d}_k]_{M, \: \cdot} \sim \mathcal{N} \left(0, K_{\mathbf{a}\mathbf{a}}^M + \sigma_M^2 \right)$. The term $K_{\mathbf{a}\mathbf{a}}^M$ denotes the Gram matrix of data points utilizing the kernel function $k^M(\cdot, \cdot)$ at the input locations $\mathbf{a}$, i.e., $[K_{\mathbf{a}\mathbf{a}}^M]_{ij} = k^M(\mathbf{a}_i, \mathbf{a}_j)$. The choice of kernel function $k^M(\cdot, \cdot)$ is shaped by prior data observations. In this study, we employed the squared exponential (SE) kernel:
\begin{ceqn}
    \begin{equation} 
        k^M({a}_i, {a}_j) = \sigma_{f, M}^2  \exp \left[-\frac{1}{2}
        \left({a}_i - {a}_j \right)^\mathsf{T} \mathbf{L}_M^{-1}
        ({a}_i - {a}_j) \right] \, , \tag{5} \label{eqn:rbf}
    \end{equation}
\end{ceqn}
Here, ${a}_i$ and ${a}_j$ represent different data points within $\mathbf{a}$, each with dimensions $n_a$-by-$1$. The hyperparameters $\sigma_{f, M}^2$ and $ \mathbf{L}_M \in \mathbb{R}^{n_a \times n_a}$ are learned from the observed dataset by maximizing the log marginal likelihood function \cite{rasmussen2006}. 

The SE kernel was selected for its smoothness and infinite differentiability, which are essential for modeling gradual changes and precise control in robotic applications \cite{brunke2021}. Its hyperparameters $\sigma_{f, M}^2$ and $\mathbf{L}_M$ enable flexible adaptation to the variability and correlation of the data. These hyperparameters enable the fine-tuning of the model to capture the underlying patterns in the data accurately, ensuring that the model is sufficiently detailed to reflect the complexities of the data without being overly sensitive to noise \cite{wang2023intuitive}. The proven effectiveness of the SE kernel in GP-based robotic control further validates its suitability to our research \cite{wang2023learning, hewing2020}. After the hyperparameters are optimized, the resulting GP model can provide predictions and uncertainty estimates for the new input points.

For the output dimension $M$, the joint distribution between $[\mathbf{d}]_{M, \: \cdot}$ and $[\mathbf{d}]_M$, which represent the observed outputs and predictions at new data points $a$, respectively,  can be formulated as
\begin{ceqn}
    \begin{equation}
        \begin{bmatrix} [\mathbf{d}]_{M, \: \cdot} \\ [\mathbf{d}]_M \end{bmatrix} \sim 
         \mathcal{N} \left( 0 , \begin{bmatrix} K_{\mathbf{a}\mathbf{a}}^M + I \sigma_ M^2 & K_{\mathbf{a}{a}}^M \\ K_{{a}\mathbf{a}}^M  &  K_{{a}{a}}^M \end{bmatrix} \right) \, , \tag{6} \label{eqn:gp_joint} 
    \end{equation}
\end{ceqn}
where $[K_{\mathbf{a}{a}}^M]_j = k^M(a_j, a)$, $K_{{a}\mathbf{a}}^M = (K_{\mathbf{a}{a}}^M)^\mathsf{T}$, and $K_{{a}{a}}^M = k^M(a,a)$. By utilizing the equations from \cite{rasmussen2006}, we can calculate the posterior distribution of $[\mathbf{d}]_M$ based on the data gathered using \eqref{eqn:gp_joint} as
\begin{ceqn}
    \begin{align} 
       \mu^M(a) &= K_{{a}\mathbf{a}}^M \left(K_{\mathbf{a}\mathbf{a}}^M + I\sigma_M^2 \right)^{-1} [\mathbf{d}]_{M, \: \cdot} \, , \tag{7a} \label{eqn:7a}   \\ 
       \Sigma^M(a) &= K_{{a}{a}}^M -  K_{{a}\mathbf{a}}^M \left(K_{\mathbf{a}\mathbf{a}}^M + I\sigma_M^2 \right)^{-1} K_{\mathbf{a}{a}}^M \, . \tag{7b} \label{eqn:7b}
    \end{align}
\end{ceqn}
The assembled GP model estimates $\mathbf{d}(\cdot)$ for the unknown function $\mathbf{g}(\cdot)$ are derived by combining each output dimension $M$ in the set $\{ 1, \cdots, n_d\}$ as follows: 
\begin{ceqn}
    \begin{align}
        &\mathbf{d}(a) \sim \mathcal{N} \left(\boldsymbol {\mu}^{\mathbf{d}}(a), \boldsymbol{\Sigma}^{\mathbf{d}}(a) \right) \, ,  \tag{8a} \label{eqn:gp_all} \\
        \boldsymbol {\mu}^{\mathbf{d}}(a) &= [\mu^1(a), \cdots, \mu^{n_d} (a)]^\mathsf{T} \in \mathbb{R}^{n_d} \, ,  \tag{8b} \label{eqn:gp_all_mean} \\
        \boldsymbol{\Sigma}^{\mathbf{d}}(a) &= \operatorname{diag}\left([\Sigma^1(a), \cdots, \Sigma^{n_d}(a)]^\mathsf{T} \right) \in \mathbb{R}^{n_d \times n_d} \, . \tag{8c} \label{eqn:gp_all_var} 
    \end{align}
\end{ceqn}

\subsection{ARX+GP model}
In our approach, a GP model is deployed to \emph{refine} the predictions of the standard ARX nominal model \eqref{eqn:arx}, resulting in an enhanced ARX+GP framework defined as follows:
\begin{ceqn}
\label{eq:arx_gp_model}
    \begin{align}
        v^H_k &= \sum_{i=1}^4 -c_i v^H_{k-i} + \sum_{i=1}^4 b_i v^{N_a}_{k-i}  = {f}(\cdot) \quad \text{ (see \ref{eqn:arx})} \, , \tag{9a} \label{eqn:arxgp_a}\\
        \tilde{v}^{H}_{k}&=\overbrace{v^H_k}^{\text {ARX prediction}}+\overbrace{{g}(v^{H}_{k-1}, v^{N_a}_{k-1})} ^{\text {GP-based correction}} \, . \tag{9b} \label{eqn:arxgp_b}
    \end{align}
\end{ceqn}
In this model, $\tilde{v}^{H}_{k}$ represents the GP-compensated velocity prediction of the HV. The ARX nominal model, denoted by ${f}(\cdot)$, is characterized by the constants $b_i$ and $c_i$. The GP model ${g}(\cdot)$ learns the divergence between the actual system behaviors and the ARX predictions. It considers both $v^{H}$ (HV velocity) and $v^{N_a}$ (velocity of the last AV). This ensures that the GP model adequately captures the dynamics of the HV as influenced by the actions of the leading AV.

To identify the system discrepancies, we use equation \eqref{eqn:arxgp_b} as well as past velocity measurements $v^{H}{j}$ and $v^{N_a}{j}$, resulting in
\begin{equation}
    \hat{v}^{H}_{j} -{v}^{H}_{j} = {g}(\mathbf{a}_{j-1}) \, ,  \tag{10} \label{eqn:gp_data_gen}
\end{equation}
where $\mathbf{a}_{j-1}$ represents the discrepancy state $(v^{H}_{j-1}, v^{N_a}_{j-1})$. Both $\hat{v}^{H}_{j}$ and $v^H_k$ refer to velocities ascertained from the collected data and predicted by the ARX model \eqref{eqn:arxgp_a}, respectively. The GP model ${g}(\cdot)$ captures the correlation between these discrepancy states and the corresponding discrepancies ${d}_{j-1}$.

In each experiment, all input--output data (one-dimensional) pairs were collected at each time step. These data points were then prepared according to equation \eqref{eqn:gp_data_gen} to create a discrepancy data set, represented as $\mathcal{D}_n = {\mathbf{a} = [\mathbf{a}_1, \cdots, \mathbf{a}_{j-1}, \cdots, \mathbf{a}_n]^\mathsf{T}, \mathbf{g} = [{g}_1, \cdots, {g}_{j-1}, \cdots, {g}_n]^\mathsf{T} }$, where $n$ is the total number of time steps in an experiment. During the GP model training phase, we optimized the kernel function hyperparameters by maximizing the log marginal likelihood of the observed disturbance data \cite{rasmussen2006, wang2023intuitive}.

\subsection{Modeling of human-driven vehicles}
\label{sec:HV_model}

\begin{figure}
    \centering
    \vspace{0.2cm}
    \includegraphics[trim=0cm 0cm 0cm 0cm, width=0.9\columnwidth]{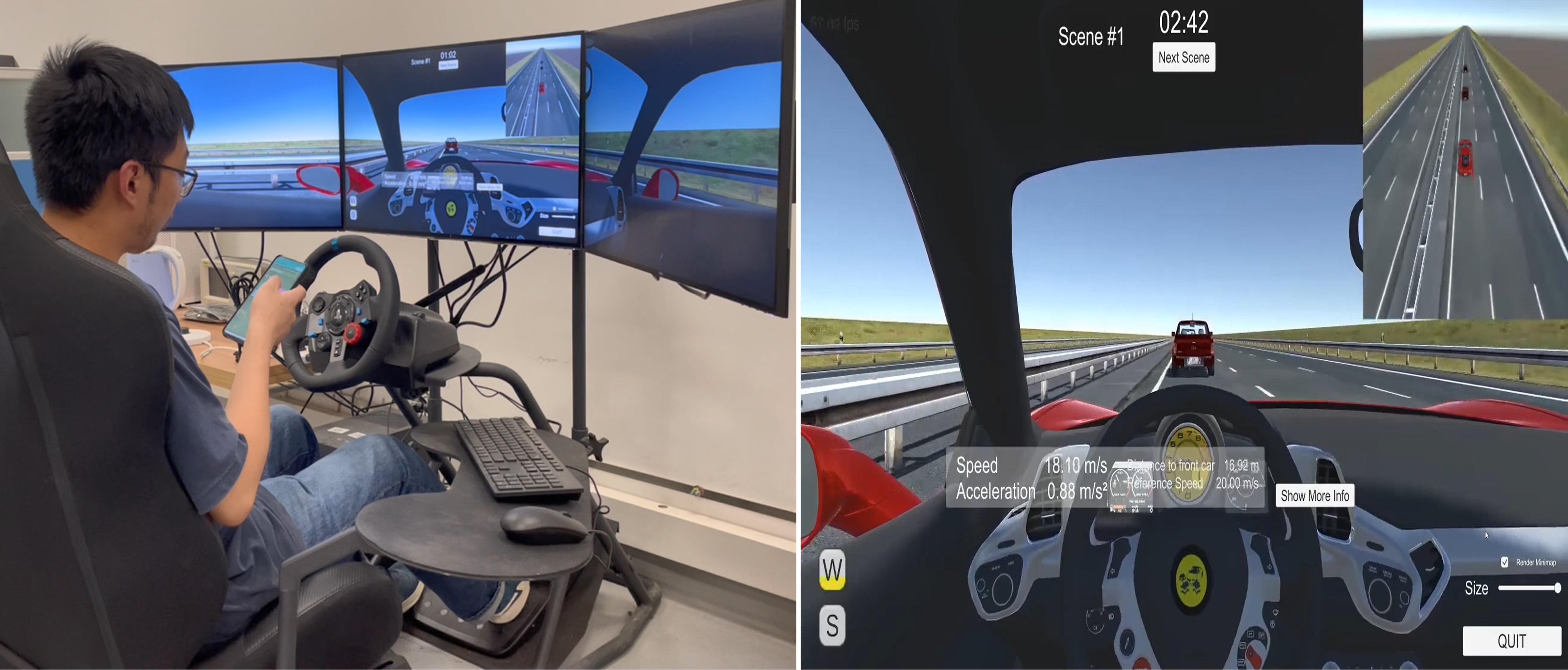}
    \caption{One of three distracted drivers in a controlled experiment, designed to gather data for estimating the HV model in a Unity driving simulator. }
    \label{figure:car_simulator}
\end{figure}

The data for estimating the HV model were gathered from a set of unique driving scenarios. Three drivers were recorded as they followed an AV platoon in a Unity simulator, as shown in Fig. \ref{figure:car_simulator}). In each experiment, the drivers were distracted by requests to answer multiple-choice algebraic questions. This distraction was introduced to simulate the challenges of distracted driving, which is particularly relevant for understanding and ensuring safe control in AV--HV interactions.

The process of data collection and the experimental setup are detailed in \cite[Sec.\ V.B]{pirani2022}. After collection, the data were processed to compute mean values for each recorded data point. These averages then informed the identification of the transfer function in equation \eqref{eqn:tf} as
\begin{equation} 
    G_{h}(s) =\frac{1+6.96 s}{1+2(0.65)(4.76) s+(4.76)^{2} s^{2}} e^{-0.512 s} \, . \tag{11} \label{eqn:tf_s}
\end{equation}
The parameters of the ARX nominal model \eqref{eqn:arx} were identified as $c_1 = -3.0227$, $c_2 = 3.3543$, $c_3 = -1.6329$, $c_4 = 0.3014$, $b_1 = 0.0063$, $b_2 = -0.0303$, $b_3 = 0.0495$, and $b_4 = -0.0254$. These values were established by restructuring \eqref{eqn:tf_s} according to \eqref{eqn:tf_z} and \eqref{eqn:arx}.

A GP model was trained using a 20\% sample from six datasets to assess the gap between the nominal model \eqref{eqn:arx} and actual human vehicle behavior. The remaining datasets were reserved for testing. Our decision to limit the size of the training dataset was driven by two main considerations. First, training the GP model with a larger dataset would incur significant computational costs. This would adversely affect the efficiency of the ARX+GP model when integrated into the MPC's prediction cycle because it would require more processing time for each prediction. Second, the nature of the collected data influenced our choice. Most of the data were recorded at speeds of 10, 15, and 20 m/s, resulting in a limited variety of driving conditions. Our tests indicated that expanding the dataset size would not result in notable improvements in the performance of the GP models. This was attributed to the lack of diversity in the speed ranges within the collected data, which limited the potential benefits of a larger training set.

\begin{figure}
    \centering
    \vspace{0.2cm}
    \includegraphics[trim=0cm 0cm 0cm 0cm, width=\columnwidth]{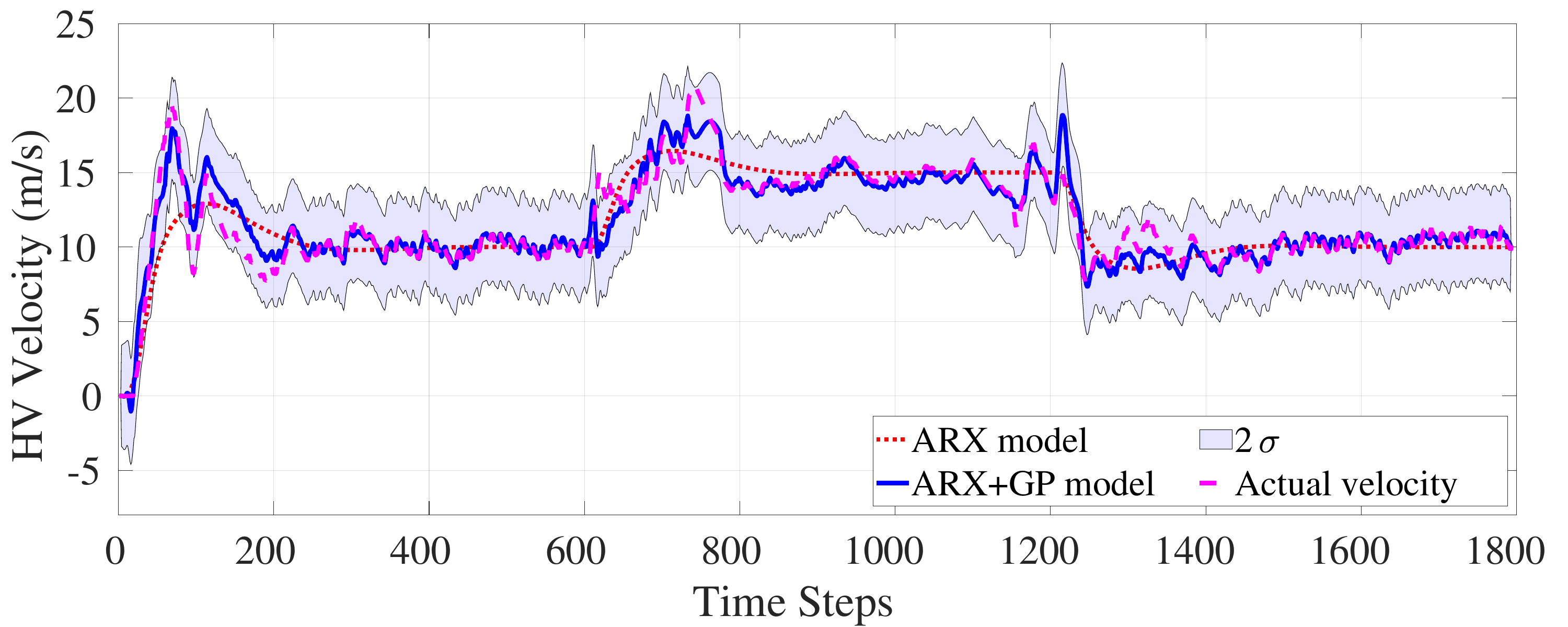}
    \caption{Performance of the ARX+GP model evaluated on one testing dataset. The HV velocity predictions of the ARX and ARX+GP models were compared by plotting them alongside the measured velocities and twice the standard deviation (2$\sigma$) determined using the GP model. The results highlight that the ARX+GP model significantly enhanced the fit of velocity curves compared with the standalone ARX model. On average, the ARX+GP model improved modeling accuracy by 35.64\% in terms of the RMSE compared with the ARX model.}
    \label{figure:GP_modeling}
\end{figure}

We tested the trained GP model using three test datasets and plotted the results for one test. Figure \ref{figure:GP_modeling} plots the predicted velocities from the ARX and ARX+GP models, along with twice the standard deviation (2$\sigma$), as assessed using the GP model. We also plotted the measured velocities of the HV to demonstrate the enhanced accuracy of the GP model.
Furthermore, we quantified the improvement in the accuracy of the GP model by computing the root mean square error (RMSE) for each test dataset. Lower RMSE values indicate a more accurate model because the RMSE measures the average deviation of the model predictions for each test dataset. We computed the mean RMSE values for both the ARX and GP+ARX models over the three test datasets. The ARX model exhibited an average RMSE of 1.88, in contrast to the ARX+GP model, which reported an average RMSE of 1.21. This outcome signified an improvement of approximately 35.64\% in accuracy. This enhancement in accuracy is clearly observed in Fig. \ref{figure:GP_modeling}, where the ARX+GP model corresponds more closely with the actual velocity data than the ARX model.

In this study, data were collected from three drivers for two reasons. First, the primary focus of our research was on modeling HV uncertainty in longitudinal car-following scenarios to improve the AV platoon control in mixed traffic. The three drivers were subjected to different driving scenarios while being distracted, which provided a concentrated and diverse dataset that reflected typical distracted driving behaviors. This method efficiently captures a range of human responses, corresponding closely with real-world longitudinal car-following conditions. Second, our predictive modeling approach, which integrates a GP model within an ARX framework, requires a balance between data sufficiency for accuracy and computational feasibility. The data from the three drivers provided a solid foundation for capturing critical behavioral patterns while maintaining computational efficiency. The diversity of these data proved adequate for training and testing our model, as shown by the enhanced accuracy of the ARX+GP model compared with the use of the ARX model alone, particularly in terms of RMSE values shown in this section. In future research, we will broaden our dataset to include more drivers and various driving conditions beyond longitudinal car-following scenarios. This expansion is expected to further enhance the accuracy of the model by capturing a wider array of human driving behaviors in diverse traffic scenarios. We also plan to develop more computationally efficient methodologies such as sparse GP models to increase the practicality and adaptability of our solutions to diverse real-world traffic scenarios.

\subsection{Comparative analysis with traditional car-following models}
As shown in Sec. \ref{sec:nominal_modeling}, human reaction time is a critical factor in modeling human driver behaviors, which is similar to its role in conventional car-following models surveyed in Sec. \ref{sec:related_work}. While it considers human reaction time, our approach diverges by incorporating a GP-learning-based component to account for the variability in human responses and environmental factors, which are not explicitly modeled in traditional car-following models. This GP component enables our model to adaptively learn from data, providing a more nuanced and data-driven representation of human-driven vehicle behavior. Although the foundational principle of our model corresponds with traditional models in terms of focusing on driver behavior and vehicle interactions, we expand this by applying machine learning to effectively manage the inherent uncertainties and variabilities in human driving. This aspect is often underrepresented in conventional car-following models. Furthermore, the integration of a first-principles transfer function with the GP model enables a more comprehensive representation of human--vehicle interaction dynamics compared with the more deterministic formulations typically observed in car-following models \cite{zhao2023safety}. Thus, our approach represents an advancement in the car-following paradigm, enhanced by machine-learning techniques, to reflect the complexities of human driving behavior more accurately.

\section{Controller Design}
\label{sec:control}

When designing control mechanisms for a fleet of AVs, the key objective is to ensure a safe distance between vehicles and maintain a reference velocity designated by the lead vehicle. This practice is generally referred to as cooperative adaptive cruise, as introduced in Sec. \ref{sec:related_work}. However, in a mixed fleet that includes both AVs and HVs, achieving reliable fleet control is more complicated, given that direct control over human-driven vehicles is not feasible. This section describes the development of a GP learning-based MPC strategy for longitudinal tracking in a mixed-vehicle fleet, considering crucial constraints, including acceleration, speed, and safe distancing. A unique aspect of this strategy is the integration of the uncertainties predicted by the GP models developed for HV modeling. This integration is particularly beneficial for enhancing safety in complex traffic scenarios, including scenarios requiring emergency braking. Our simulation studies highlighted the effectiveness of the proposed learning-based MPC. Compared with the baseline MPC, which relies solely on the nominal ARX model, the learning-based approach has superior performance. It achieves higher travel speeds for both AVs and HVs while providing more robust safety guarantees. This is particularly important in mixed-fleet scenarios, where the behavior of HVs can vary significantly and unpredictably, thereby requiring a more adaptable and responsive control strategy.

\subsection{System model}
To maintain a safe distance within the mixed platoon of vehicles, we establish a distance constraint between AVs, defined by a constant $\Delta$, such that $p_{k}^{\mathbf{n_a}-1}-p_{k}^{\mathbf{n_a}} > \Delta$. However, owing to the uncertainty of the HV behavior model, a probabilistic constraint, or chance constraint, is introduced to ensure a safe distance between the trailing AV and HV. This constraint is expressed as
\begin{ceqn}
    \begin{align}  
    v_{k+1}^{\mathbf{n_a}} &= v_{k}^{\mathbf{n_a}} + T \, \mathrm{acc}_{k}^{\mathbf{n_a}} \, , \tag{12a} \label{eqn:av_a} \\
    p_{k+1}^{\mathbf{n_a}} &= p_{k}^{\mathbf{n_a}} + T \, v_{k}^{\mathbf{n_a}}  \, . \tag{12b} \label{eqn:av_b}
    \end{align}
\end{ceqn}
where $0<T \ll 1$ indicates the sample time, and $\mathrm{acc}_{k}^{\mathbf{n_a}}$ denotes the acceleration of $A^{\mathbf{n_a}}$. AVs are assumed to be deterministic, with their states measured and communicated error-free, i.e., $\Sigma(v_{k}^\mathbf{n_a}) = 0$. By applying \eqref{eqn:arxgp_b}, the HV model is derived as follows:
\begin{ceqn} 
    \begin{align} 
        \tilde{v}_{k}^{H} &= v_{k}^{H} + {d}(v_{k-1}^{H}, {v}_{k-1}^{N_a}) \, , \tag{13a} \label{eqn:sys_model_a} \\
        p_{k+1}^{H} &= p_{k}^{H} + T \, \tilde{v}_{k}^{H} \, , \nonumber \\ 
        &= p_{k}^{H} + T \, v_{k}^{H} + T \, {d}(v_{k-1}^{H}, {v}_{k-1}^{N_a}) \, . \tag{13b} \label{eqn:sys_model_b}
    \end{align}
\end{ceqn}
In this context, $\tilde{v}_{k}^{H}$ represents the compensated HV velocity obtained from the GP model, whereas $v_{k}^{H}$ (computed using \eqref{eqn:arx}) and $p_{k}^{H}$ denote the velocity and position states, respectively, at the current time step $k$. At $k$, both the variances $\Sigma(v_{k}^{H})$ and $\Sigma(p_{k}^{H})$ are zero. 

Using \eqref{eqn:sys_model_b} and \eqref{eqn:gp_all}, we derive the propagation of the HV position mean $\mu \left(p_{k+1}^{H} \right)$ as follows:
\begin{ceqn} 
    \begin{align} 
        \mu \left(p_{k+1}^{H} \right) &= \mu \left( p_{k}^{H} + T \, v_{k}^{H} + T \, {d}(v_{k-1}^{H}, {v}_{k-1}^{N_a}) \right) \, , \label{eqn:vel_prop_a} \tag{14a} \\
        &= \mu \left( p_{k}^{H} \right) + T \, \mu \left( v_{k}^{H} \right) + T \, \mu \left( {d}(v_{k-1}^{H}, {v}_{k-1}^{N_a}) \right) \, , \label{eqn:vel_prop_b} \tag{14b} \\
        &= \mu \left( p_{k}^{H} \right) + T \, \mu \left( v_{k}^{H} \right) + T \, \mu^d \left( v_{k-1}^{H}, {v}_{k-1}^{N_a} \right) \, . \label{eqn:vel_prop_c} \tag{14c}
    \end{align}
\end{ceqn}
To express the equation more concisely, we denote the HV position mean at time step $k+1$ on the left side of \eqref{eqn:vel_prop_c} as $\mu_{k+1}^{p^{H}}$; hence, we can rewrite the terms on the right-hand side of the equation. Thus, equation \eqref{eqn:vel_prop_c} is compactly represented as
\begin{ceqn} 
    \begin{align} 
        \mu_{k+1}^{p^{H}} &= \mu_{k}^{p^{H}} + T \, v_{k}^{H} + T \, \mu^d \left( v_{k-1}^{H}, {v}_{k-1}^{N_a} \right), \label{eqn:vel_prop_d} \tag{14d}
    \end{align}
\end{ceqn}
with the initial value $\mu_{0}^{p^{H}} = p_{k}^{H}$. Using \eqref{eqn:sys_model_b} and \eqref{eqn:gp_all}, we derive the propagation of the HV position variance $\Sigma \left(p_{k+1}^{H} \right)$ as:
\begin{ceqn} 
    \begin{align} 
        \Sigma \left(p_{k+1}^{H} \right) &= \Sigma \left( p_{k}^{H} + T \, v_{k}^{H} + T \, {d}(v_{k-1}^{H}, {v}_{k-1}^{N_a}) \right) \, , \tag{15a} \label{eqn:15a} \\
        & = \Sigma(p_{k}^{H}) + \Sigma(T \, v_{k}^{H}) + \Sigma \left( T \, {d}(v_{k-1}^{H}, {v}_{k-1}^{N_a}) \right) \, , \tag{15b} \label{eqn:15b} \\
        & = \Sigma (p_{k}^{H}) + T^2 \Sigma^d \left(v_{k-1}^{H}, {v}_{k-1}^{N_a}\right) \, . \tag{15c} \label{eqn:15c} 
    \end{align}
\end{ceqn}
Similarly, to express the equation more succinctly, we denote the HV position variance at time step $k+1$ on the left-hand side of \eqref{eqn:15c} as $\Sigma_{k+1}^{p^{H}}$; hence, we rewrite the terms on the right-hand side of the equation. This enables us to compactly represent equation \eqref{eqn:15c} as
\begin{ceqn} 
    \begin{align} 
        \Sigma_{k+1}^{p^{H}} = \Sigma_{k}^{p^{H}} + T^2 \Sigma^d \left(v_{k-1}^{H}, {v}_{k-1}^{N_a}\right) \ . \tag{15d} \label{eqn:HV_var}
    \end{align}
\end{ceqn}
Here, the initial value is $\Sigma \left(p_{0}^{H} \right) = 0$. Note that the HV variance propagation does not consider the covariance between $p_{k}^{H}$ and $v_{k}^{H}$. 

\subsection{Safe distance chance constraint}

For a mixed-vehicle platoon comprising AVs, denoted by $A^{\mathbf{n_a}}$, a single HV is represented by $H$. Here, $\mathbf{n_a} = \{1, 2, \cdots, N_a\}$, where $N_a$ denotes the number of AVs at $k$, the velocity and position of $A^{\mathbf{n_a}}$ are denoted by $p^{\mathbf{n_a}}_k$ and $v^{\mathbf{n_a}}_k$, respectively. The kinematic model of AVs is expressed as follows:
\begin{equation} 
    \mathrm{Pr}\left(p_{k}^{N_a}-(p_{k}^{H} + \Delta) > \Delta_\text{ext} \right) \geq p_{\text{def}} \, . \tag{16} \label{eqn:chance_constraint} 
\end{equation}
In this context, $\Delta$ denotes the predetermined safe distance between AVs, whereas $\Delta_\text{ext} \geq 0$ is an additional distance incorporated to accommodate the stochastic behavior of the HV. The desired satisfaction level is denoted by $p_{\text{def}}$. The distance constraint $\mathcal{X}$ can be expressed as a single half-space constraint $\mathcal{X}^{hs} := \bigl\{x \vert h^{\top}x \leq b \bigl\}$, $ h \in \mathbb{R}^n$, $b \in \mathbb{R}$, and 
\begin{equation} 
    \begin{bmatrix} -1 & 1\end{bmatrix} \begin{bmatrix} p_{k}^{N_a} & p_{k}^{H}+\Delta \end{bmatrix}^{\top} \leq -\Delta_\text{ext} \, , \tag{17} \label{eqn:17} 
\end{equation}
where $h^{\top} = \begin{bmatrix} -1 & 1\end{bmatrix}$, $x := \begin{bmatrix} p_{k}^{N_a} & p_{k}^{H}+\Delta \end{bmatrix}^{\top}$, and $b = -\Delta_\text{ext}$. In \cite{hewing2019}, a method to tighten the constraint on the state mean was obtained as
\begin{equation} 
    \mathcal{X}^{h s}\left(\Sigma_{i}^{x}\right):=\left\{x \mid h^{\top} x \leq b-\phi^{-1}\left(p_{\text{def}}\right) \sqrt{h^{\top} \Sigma_{i}^{x} h}\right\} \, , \tag{18} \label{eqn:18} 
\end{equation}
where $\phi^{-1}$ is the inverse of the cumulative distribution function (CDF). In our scenario, 
\begin{equation} 
    \Sigma_{k}^{x} := \begin{bmatrix} \Sigma_{k}^{p^{N_a}} \\ \Sigma_{k}^{p^{H}}+\Delta \end{bmatrix} = \begin{bmatrix} 0 & 0 \\ 0 & \Sigma_{k}^{p^{H}}\end{bmatrix} \, . \tag{19} \label{eqn:19} 
\end{equation}
In this scenario, no correlation exists between the position of an HV and the leading AV, denoted by $\Sigma_{k}^{p^{N_a}} = 0$. The position variance of the HV, denoted by $\Sigma_{k}^{p^{H}}$, is computed using \eqref{eqn:HV_var}. To achieve a more ``tightened'' constraint on the position state, we substitute \eqref{eqn:17} and \eqref{eqn:19} into \eqref{eqn:18} to derive the following expression:
\begin{equation} 
    p_{k}^{N_a}-p_{k}^{H} \geq \Delta + \Delta_\text{ext} + \phi^{-1}\left(p_{\text{def}}\right) \sqrt{ \Sigma_{k}^{p^{H}}} \, . \tag{20} \label{eqn:chance_constraint_final} 
\end{equation}

The safe distance probability constraint presented in \eqref{eqn:chance_constraint} is transformed into a deterministic equation, as described in \eqref{eqn:chance_constraint_final}. When a high level of constraint satisfaction assurance is desired, the additional term $\Delta_\text{ext}$ can be set to zero, which results in further simplification. The AV--HV safe distance is adaptively adjusted, utilizing the uncertainty estimates of the HV from the GP model, as indicated in \eqref{eqn:HV_var}. This adjustment ensures that the safe distance requirement always exceeds $\Delta$ under all the conditions.

\subsection{GP learning-based MPC}
\label{sec:GP-MPC}

We developed an MPC strategy that integrates the proposed modeling method for HVs within a mixed-vehicle platoon scenario. The platoon consists of $N_a$ AVs and one HV, as shown in Figure \ref{figure:1}. This strategy can be developed as follows:
\begin{ceqn}
    \begin{align} 
       \underset{\mathbb{V}}{\text{min}} \sum_{i=k}^{k+N} \Big\| v &_{{i+1}|k}^1 - v_{{i+1}|k}^\text{ref} \Big\|^2_{Q_1} \nonumber + \sum_{\mathbf{n_a}=2}^{N_a}\sum_{i=k}^{k+N} \Big\| v_{{i+1}|k}^\mathbf{n_a} - v_{{i+1}|k}^{\mathbf{n_a}-1} \Big\|^2_{Q_2}  \nonumber \\ & \qquad \qquad \qquad \quad + \sum_{\mathbf{n_a}=1}^{N_a}\sum_{i=k}^{k+N-1} \Big\| \mathrm{acc}_{{i}|k}^{\mathbf{n_a}} \Big\|^2_R  \tag{21a} \label{eqn:mpc_a} \\
       \text{with} \ \mathbb{V} = & \left\{v_{{i}|k}^1, v_{{i}|k}^\mathbf{n_a}, v_{{i}|k}^{H}, p_{{i}|k}^\mathbf{n_a}, \mu_{{i}|k}^{p^{H}}, \Sigma_{{i}|k}^{p^{H}}, \mathrm{acc}_{{i}|k}^{\mathbf{n_a}} \right\} \, \nonumber \\
        \text {subject to} \nonumber \\
        v_{{i+1}|k}^\mathbf{n_a} &= v_{{i}|k}^\mathbf{n_a} + T \, \mathrm{a}_{{i}|k}^\mathbf{n_a} \, , \quad \mathbf{n_a} = \{1, 2, \cdots, N_a\} \tag{21b} \label{eqn:mpc_b}\\ 
        p_{{i+1}|k}^\mathbf{n_a} &= p_{{i}|k}^\mathbf{n_a} + T \, v_{{i}|k}^\mathbf{n_a} \, , \quad \mathbf{n_a} = \{1, 2, \cdots, N_a\} \tag{21c} \label{eqn:mpc_c}\\
        v^{H}_{{i}|k} &= {f}\left(v_{i-1:i-4 | k}^{H}, {v}_{i-1:i-4 | k}^{N_a} \right) \quad \text{ (see \ref{eqn:arx})} \, , \tag{21d} \label{eqn:mpc_d}\\
        \mu_{{i+1}|k}^{p^{H}} &= \mu_{{i}|k}^{p^{H}} + T \, v_{{i}|k}^{H} + T \, \mu^d (v_{{i-1}|{k}}^{H}, {v}_{{i-1}|{k}}^{N_a}) \, , \tag{21e} \label{eqn:mpc_e}\\
        \Sigma_{{i+1}|k}^{p^{H}} &= \Sigma_{{i}|k}^{p^{H}} + T^2 \Sigma^d (v_{{i-1}|{k}}^{H}, {v}_{{i-1}|{k}}^{N_a}) \, , \tag{21f} \label{eqn:mpc_f}\\
        p_{{i}|k}^{\mathbf{n_a}-1} & - p_{{i}|k}^\mathbf{n_a} \geq \Delta \, , \quad \mathbf{n_a} = \{2, \cdots, N_a\}\tag{21g} \label{eqn:mpc_g}\\
        p_{{i}|k}^{N_a} & - \mu_{{i}|k}^{p^{H}} \geq \Delta + \phi^{-1}\left(p_{\text{def}}\right) \sqrt{ \Sigma_{{i}|k}^{p^{H}}} \, , \tag{21h} \label{eqn:mpc_h}\\
        v_{\text{min}} & \leq v_{{i}|k}^\mathbf{n_a} \leq v_{\text{max}} \, , \quad \mathbf{n_a} = \{1, 2, \cdots, N_a\} \tag{21i} \label{eqn:mpc_i}\\
        \mathrm{acc}_{\text{min}} & \leq \mathrm{acc}_{{i}|k}^{\mathbf{n_a}} \leq \mathrm{acc}_{\text{max}} \, , \quad \mathbf{n_a} = \{1, 2, \cdots, N_a\} \tag{21j} \label{eqn:mpc_j}\\
        v_{{0}|{k}}^1 & = v_{{k}|{k}}^1 \, ,  v_{{0}|{k}}^\mathbf{n_a} = v_{{k}|{k}}^\mathbf{n_a} \, , v_{{0}|{k}}^{H} = v_{{k}|{k}}^{H} \, , \tag{21k} \label{eqn:mpc_k}\\
        v_{{-1}|{k}}^\mathbf{n_a} &= v_{{k-1}|{k}}^\mathbf{n_a} \, , \ v_{{-2}|{k}}^\mathbf{n_a} = v_{{k-2}|{k}}^\mathbf{n_a} \, , \nonumber \\ 
        v_{{-3}|{k}}^\mathbf{n_a} &= v_{{k-3}|{k}}^\mathbf{n_a} \, , \ v_{{-4}|{k}}^\mathbf{n_a} = v_{{k-4}|{k}}^\mathbf{n_a} \, , \tag{21l} \label{eqn:mpc_l}\\ 
        v_{{-1}|{k}}^{H} &= v_{{k-1}|{k}}^{H} \, , \ v_{{-2}|{k}}^{H} = v_{{k-2}|{k}}^{H} \, , \nonumber \\ 
        v_{{-3}|{k}}^{H} &= v_{{k-3}|{k}}^{H} \, , \ v_{{-4}|{k}}^{H} = v_{{k-4}|{k}}^{H} \, , \tag{21m} \label{eqn:mpc_m}\\
        \mu_{{0}|{k}}^{p^{H}} &= p_{{k}|{k}}^{H} \, , \ \Sigma_{{0}|{k}}^{p^{H}} = 0 \ . \tag{21n} \label{eqn:mpc_n}
    \end{align}
\end{ceqn}

The current moment in the system (either hardware or simulated) is represented by variable $k$, with the process initiating at $k=0$. Within the prediction horizon of the MPC, the initial prediction step is given as $i=1$ and the variables at $i=0$ are established with measurements, as denoted on the right-hand sides of \eqref{eqn:mpc_k}, \eqref{eqn:mpc_l}, \eqref{eqn:mpc_m}, and \eqref{eqn:mpc_n}. The cost function \eqref{eqn:mpc_a} contains three positive weights, $Q_1$, $Q_2$, and $R$, which are used to balance the deviation between the desired velocity and the speed of the leading AV, discrepancies in speed between neighboring AVs, and control inputs, respectively.

The equality constraints expressed in \eqref{eqn:mpc_b}--\eqref{eqn:mpc_f} involve the AV models specified in \eqref{eqn:av_a} and \eqref{eqn:av_b}, the nominal ARX model of the HV described in \eqref{eqn:arx}, and the mean and variance propagation for the HV position deduced in \eqref{eqn:vel_prop_d} and \eqref{eqn:HV_var}. The inequality constraints include the velocity \eqref{eqn:mpc_i}, acceleration \eqref{eqn:mpc_j}, and safe distance \eqref{eqn:mpc_g}--\eqref{eqn:mpc_h} requirements. The combination of these constraints and conditions ensures safe and efficient control of the mixed-vehicle fleet under the GP-MPC strategy. To highlight the essential role of the GP models in managing the uncertainties associated with HVs, we summarize the proposed GP-MPC strategy in Algorithm \ref{alg:gp_mpc}. This algorithm represents the practical application of our GP-MPC strategy, guiding the optimal control actions for AVs at each step of the control horizon and dynamically incorporating the probabilistic nature of HV behavior through GP model estimations. The algorithm is presented as follows. 
\begin{algorithm}
\caption{GP Learning-Based MPC for Mixed-Vehicle Platoon}
\label{alg:gp_mpc}
\begin{algorithmic}[1]
\fontsize{11pt}{13pt}\selectfont
\State \textbf{Input:} Initial states of AVs and HVs, $T$, $\Delta$, $\Delta_{\text{ext}}$, $p_{\text{def}}$, $v_{\text{min}}$, $v_{\text{max}}$, $\mathrm{acc}_{\text{min}}$, $\mathrm{acc}_{\text{max}}$
\State \textbf{Output:} Optimal control actions for AVs
\For{each time step $k$ in the control horizon}
    \State \textbf{Formulate MPC Optimization Problem:}
    \State Define the cost function (21a).
    \State \textbf{Integrate Constraints:}
    \State \quad a. Velocity and position for AVs using (21b) and (21c).
    \State \quad b. \hspace{-0.09cm}Velocity and position for the HV (21d--21f), incorporating GP \phantom{for} \phantom{for} \hspace{0.2cm} model estimations to address HV uncertainties.
    \State \quad c. \ Safe distance constraints for AVs (21g) and adaptive safe distance \phantom{for} \phantom{for} \hspace{0.8cm} between the last AV and HV (21h), using GP uncertainty estimates.
    \State \quad d. Velocity limits (21i) and acceleration limits (21j) for AVs.
    \State Set initial conditions for current and previous time steps (21k--21n)
    \State \textbf{Solve MPC Optimization} to determine control actions.
    \State \textbf{Apply Control Inputs} to AVs based on optimization solution.
\EndFor
\end{algorithmic}
\end{algorithm}

\subsection{Summary}
The integration of the GP models into our MPC framework plays an essential role in managing the inherent uncertainties of HVs within a mixed platoon of AVs and HVs. The GP model captures the probabilistic nature of HV behavior, enabling us to predict and quantify the uncertainty in HV actions. This predictive capability is critical for adapting a control strategy for AVs to ensure safety and efficiency. Specifically, the GP model predicts the future positions and velocities of the HV with the associated uncertainties (\ref{eqn:sys_model_a} and \ref{eqn:sys_model_b}). These predictions are then used to formulate the chance constraints (\ref{eqn:chance_constraint}), which are integral to maintaining safe distances within a mixed platoon, particularly under dynamic and uncertain traffic conditions. 

In our control strategy design, the MPC framework (\ref{eqn:mpc_a}–-\ref{eqn:mpc_n}) explicitly integrates the GP-based position prediction (\ref{eqn:vel_prop_d}), uncertain propagation (\ref{eqn:HV_var}), and HV--AV adaptive safe distance(\ref{eqn:chance_constraint_final}) as constraints. This GP-based constraint integration ensures that our control strategy effectively accounts for the stochastic nature of HVs, thus ensuring that the control strategy remains robust and reliable, even in the presence of unpredictable human behavior. Our simulation studies, described in the following section, demonstrated that the proposed GP-MPC approach outperforms traditional control strategies, providing enhanced safety margins and operational efficiency in mixed-traffic scenarios.

\section{Simulation and Experiments}
\label{sec:simulation}

This section presents the results of synthetic trials conducted on a mixed-vehicle platoon consisting of two AVs followed by an HV. We compared the performance of the proposed GP-MPC methodology with that of a baseline MPC to demonstrate the efficacy of the proposed approach in mixed-traffic scenarios. 

\subsection{Simulation setup}
\label{sec:simulation_setup}
\textbf{Simulation Environment:} All simulations were conducted using MATLAB R2022a on an Ubuntu 20.04 system. The source code is available at: \url{https://github.com/CL2-UWaterloo/GP-MPC-of-Platooning}. 

\textbf{Initial Conditions:} Each vehicle in the platoon started from a stationary position with an initial velocity of zero. The first AV was positioned at \(p=0\), followed by the second AV at \(p=-20\) m and the HV at \(p=-40\) m.

\textbf{Simulation Parameters:} The sample time was set at \(T = 0.1\) s, and the prediction horizon for both MPC methodologies was \(N=10\). For the MPC cost function, we used \(Q_1 = Q_2 = 5\) and \(R = 10\). The acceleration limits were set to \(a_{\text{max}} = 5 \, \text{m/s}^2\) and \(a_{\text{min}} = -5 \, \text{m/s}^2\), and the velocity limits were \(v_{\text{max}} = 35 \, \text{m/s}\) and \(v_{\text{min}} = -35 \, \text{m/s}\). We defined the chance constraint satisfaction probability in \eqref{eqn:chance_constraint_final} as $p_{\text{def}}=0.95$.

\textbf{Comparison with baseline MPC:} The nominal MPC used standard HV models as defined by equations \eqref{eqn:mpc_d} and \eqref{eqn:mpc_e}, excluding the third GP component. Unlike GP-MPC, the nominal MPC does not incorporate position variance propagation, as indicated in \eqref{eqn:mpc_f}, or the second adaptive segment of the distance constraint between the HV and the trailing AV from \eqref{eqn:mpc_h}. In both the nominal MPC and GP-MPC, we simulated the HV model using the ARX+GP model, as developed in equations \eqref{eqn:sys_model_a} and \eqref{eqn:sys_model_b}.

\subsection{Simulation results}
Our simulations focused on two scenarios: constant-velocity tracking and emergency braking. These scenarios were selected to evaluate the effectiveness of the proposed GP-MPC strategy against the nominal MPC under both steady and dynamic conditions.

\begin{figure}
    \centering
    \vspace{0.05cm}
    \subfloat{{\includegraphics[trim=0cm 0cm 0cm 0cm, width=0.98\columnwidth]{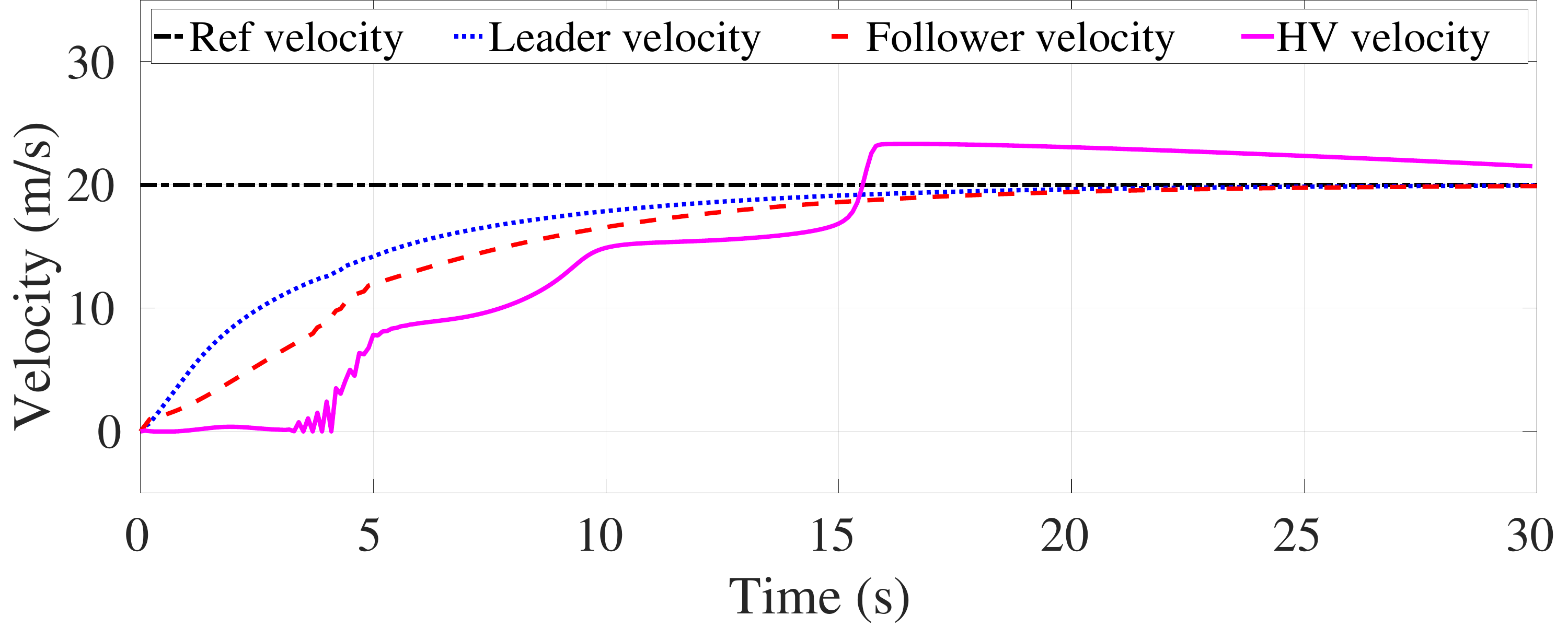} }}
    \qquad \qquad 
    \vspace{-0.6cm}
    \subfloat{{\includegraphics[trim=0cm 0cm 0cm 0.0cm, width=0.98\columnwidth]{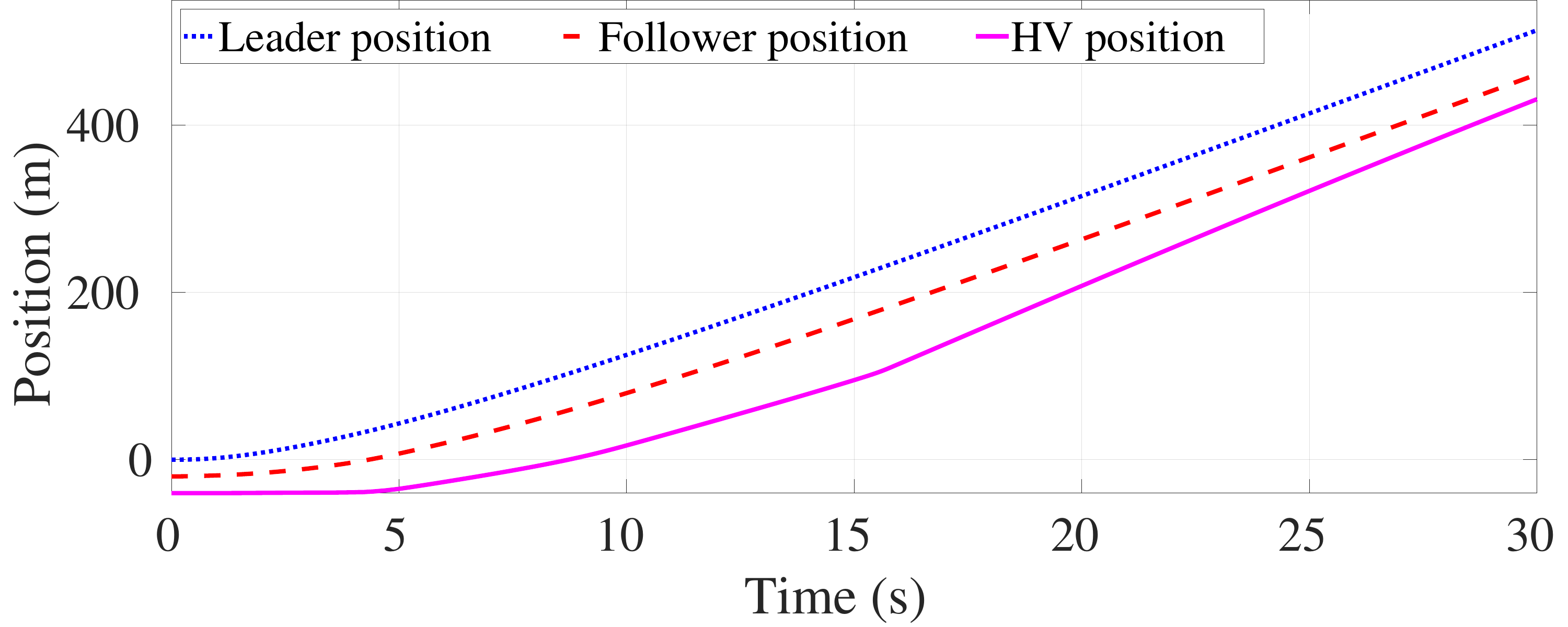} }}
    \qquad \qquad 
    \subfloat{{\includegraphics[trim=0.0cm 0cm 0.0cm 0.0cm, width=0.98\columnwidth]{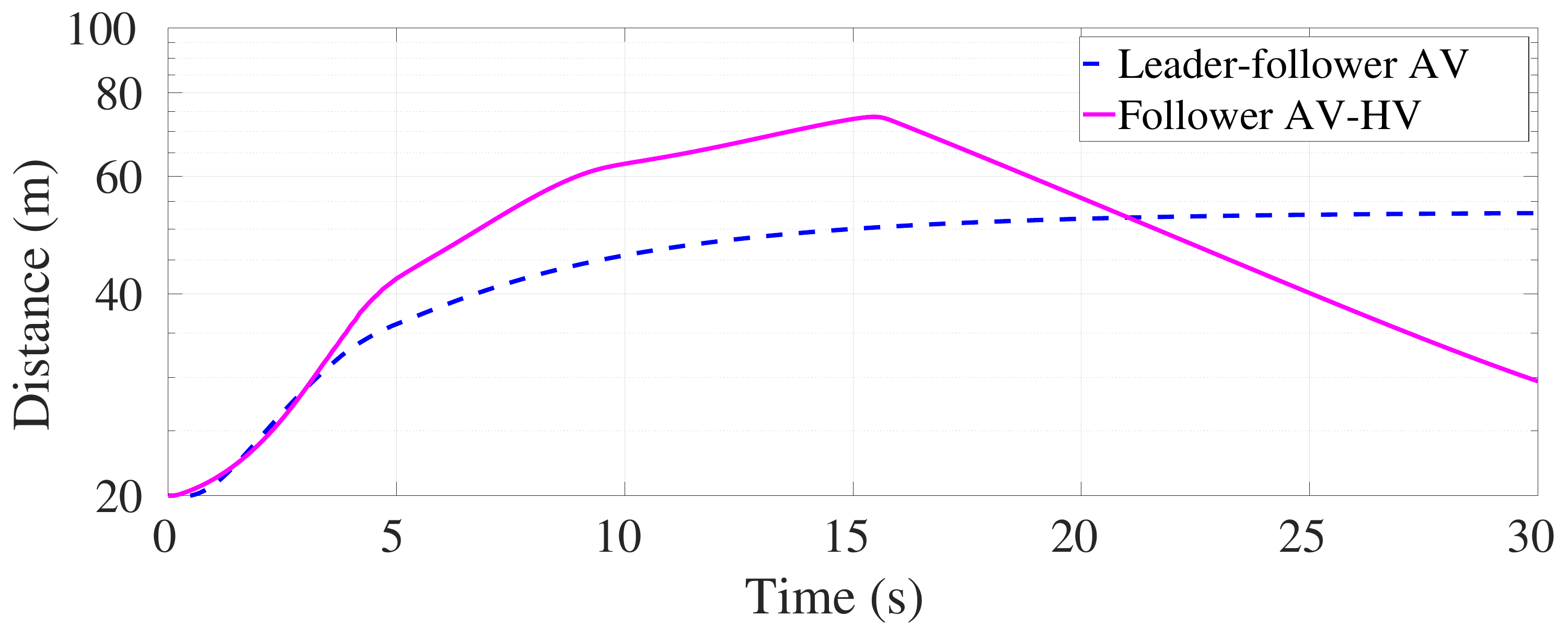} }}
    \caption{Results of a constant velocity-tracking simulation utilizing GP-MPC. Sequentially from top to bottom, the plots illustrate the velocity response, position trajectories, and inter-vehicle distance. These graphs demonstrate the stable performance of GP-MPC in achieving and maintaining the desired platoon behavior under constant velocity conditions. }
    \label{figure:gp_simulation}
\end{figure}

\textbf{Constant-Velocity Tracking:} 
In the initial trial, the leading AV was assigned a reference speed of \(v^{\text{ref}} = 20 \, \text{m/s}\). The results are shown in Fig. \ref{figure:gp_simulation}, where the velocity response, position trajectories, and inter-vehicle distance are presented in a top-down format. The top velocity response plot demonstrates that both AVs smoothly adjusted their velocities to correspond with the reference velocity set for the leading AV. The follower AV closely matched the velocity of the leading AV, demonstrating the effective convergence of the control strategy. Notably, the HV also achieved a stable velocity, indicating that the GP-MPC strategy can influence the HV behavior toward the desired traffic flow. The middle position trajectory plot shows a linear increase in position over time for all vehicles, confirming consistent movement and velocity maintenance after the initial acceleration phase. The steady trajectory slopes indicate that the vehicles maintained a uniform speed that corresponded with the reference velocity after the convergence period. In the bottom inter-vehicle distance plot, after an initial period of adjustment, the distance between the AVs stabilized and became constant after 20 s. This suggested that the platoon reached a steady state where the distances between vehicles were maintained, thereby reflecting successful platooning control. 

The top graph shows a disturbance in the HV velocity between 4 and 5 s. This deviation from the expected behavior is attributed to the limited training data diversity of the GP model, as detailed in Section \ref{sec:HV_model}. During this period, GP-MPC was unable to compensate for the discrepancy when the HV velocity was between 0 and 10\,\text{m/s}. However, the system managed to recover and align the HV's velocity with the reference velocity, demonstrating the resilience of the GP-MPC strategy. 

To ensure clarity, we do not include the nominal MPC plots as they were very similar to those of GP-MPC. The focus here is to demonstrate the stable performance of GP-MPC in achieving and maintaining the desired platoon behavior under constant-velocity conditions. The results confirmed that GP-MPC is effective in guiding a platoon into a stable traffic flow, even in the presence of minor disturbances. Compared with the nominal MPC, GP-MPC exhibited a modest improvement in maintaining a larger minimum distance between the HV and trailing AV. As illustrated in Tab. \ref{tab:simulation_metrics_con}, the minimum distance under GP-MPC was 29.64 m, slightly higher than the 29.51 m achieved by the nominal MPC. This improvement demonstrates the ability of GP-MPC to consider the uncertainties inherent in HV behaviors. In addition, all the vehicles within the mixed platoon achieved higher speeds when using this strategy. 
\begin{table}
    \caption{Constant reference velocity tracking simulation results: The location of each vehicle and the minimum relative distance of the HV--AV platoon.
    }
    \label{tab:simulation_metrics_con}
    \begin{center}
        \begin{tabular}{|c|c|c|c|c|}
            \hline Controller & AV 1 & AV 2 & HV & Min HV--AV  \\
            \hline Nominal MPC & 508.62 m & 452.84 m & 423.33 m & 29.51 m\\
            \hline GP-MPC & \textbf{513.81} m & \textbf{460.97} m &  \textbf{431.33} m & \textbf{29.64} m\\
            \hline
        \end{tabular}
    \end{center}
\end{table}

\begin{figure}
    \centering
    \vspace{0.05cm}
    \subfloat{{\includegraphics[trim=0cm 0cm 0cm 0cm, width=0.94\columnwidth]{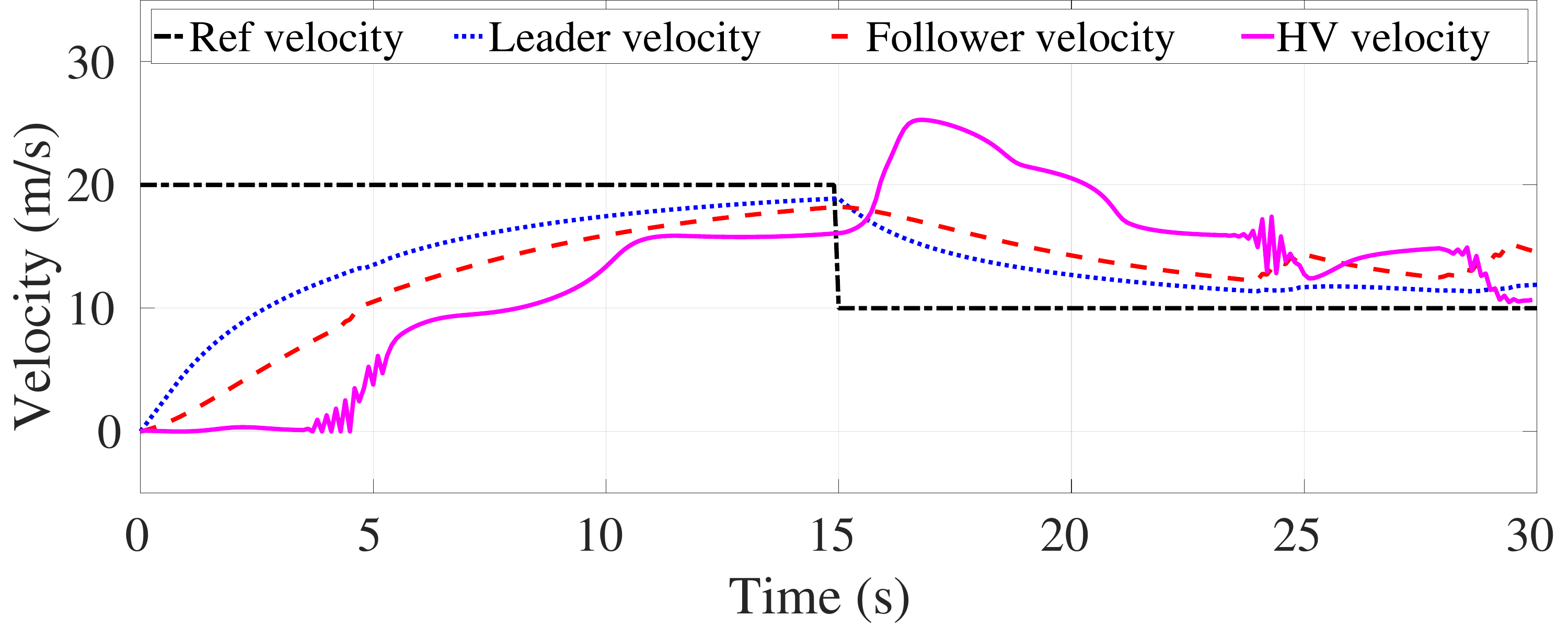} }}
    \qquad \qquad 
        
    \vspace{-0.8cm}
    \subfloat{{\includegraphics[trim=0cm 0cm 0cm 0.0cm, width=0.97\columnwidth]{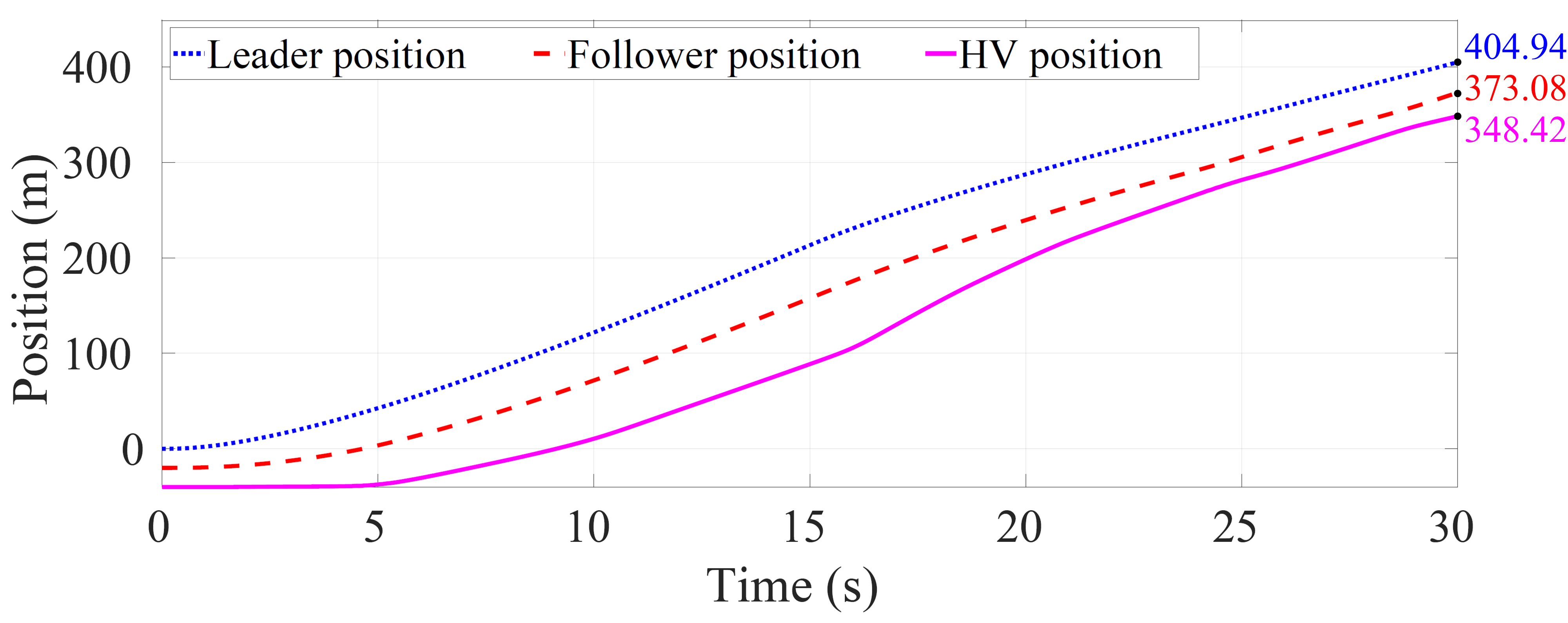} }}
    \qquad \qquad 
    \subfloat{{\includegraphics[trim=0.0cm 0cm 0.0cm 0.0cm, width=0.94\columnwidth]{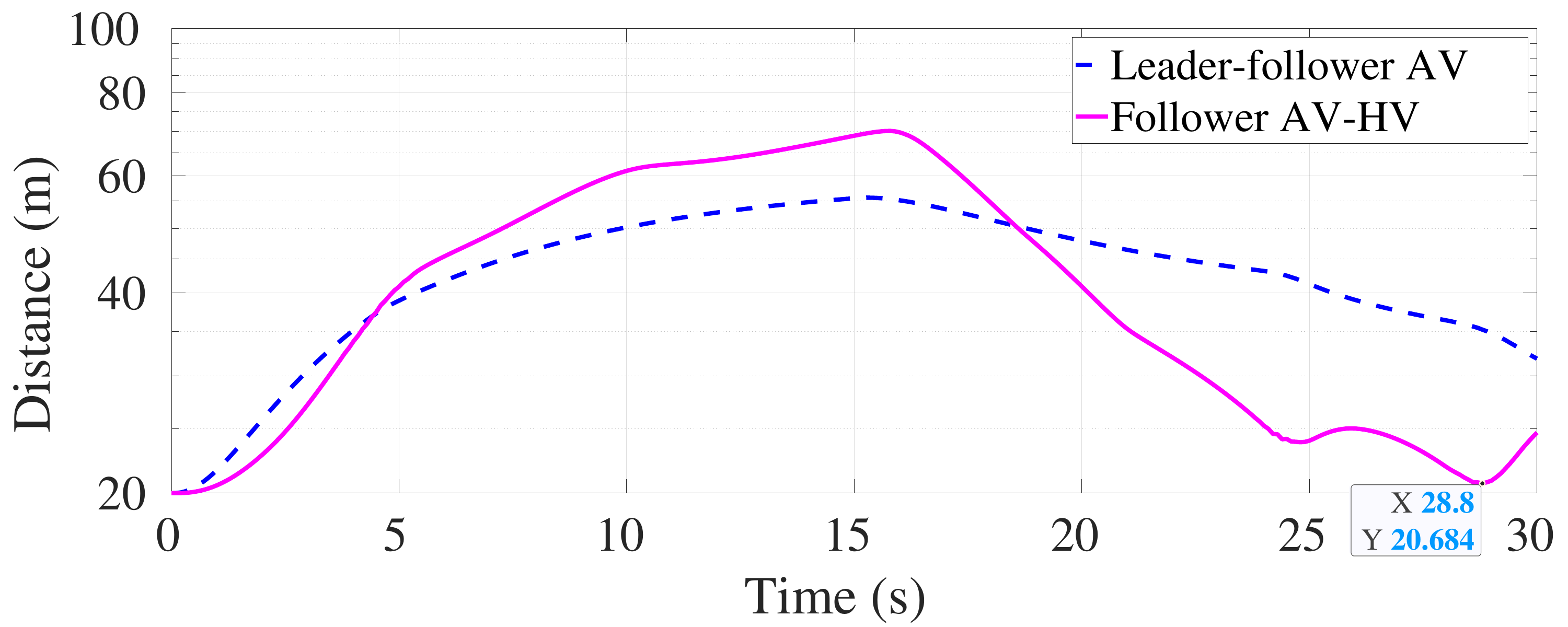} }}
    \caption{Results of an emergency braking simulation using the baseline MPC. The figures represent, in descending order, the velocity response, position trajectories, and inter-vehicle distance. } 
\label{figure:nominal_simulation_braking}
\end{figure}

\begin{figure}
    \centering
    \vspace{0.05cm}
    \subfloat{{\includegraphics[trim=0cm 0cm 0cm 0cm, width=0.94\columnwidth]{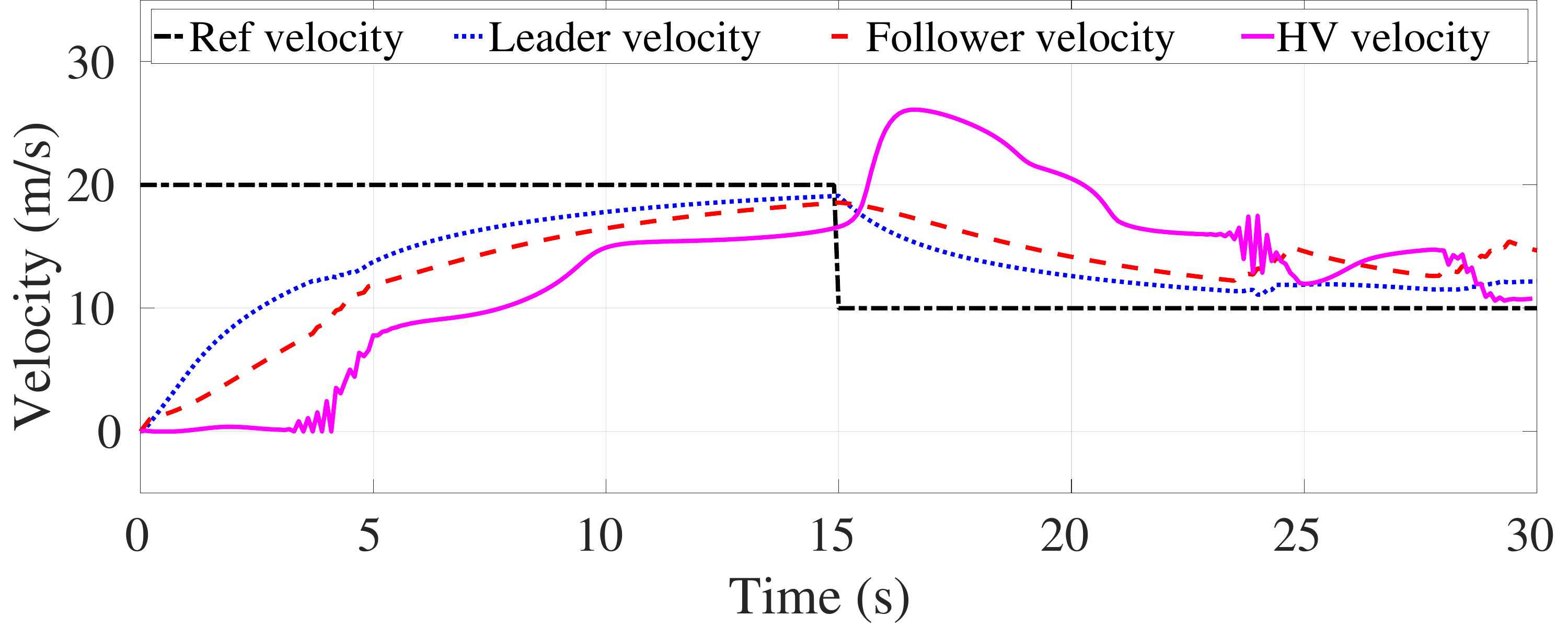} }}
    \qquad \qquad 
        
    \vspace{-0.8cm}
    \subfloat{{\includegraphics[trim=0cm 0cm 0cm 0.0cm, width=0.97\columnwidth]{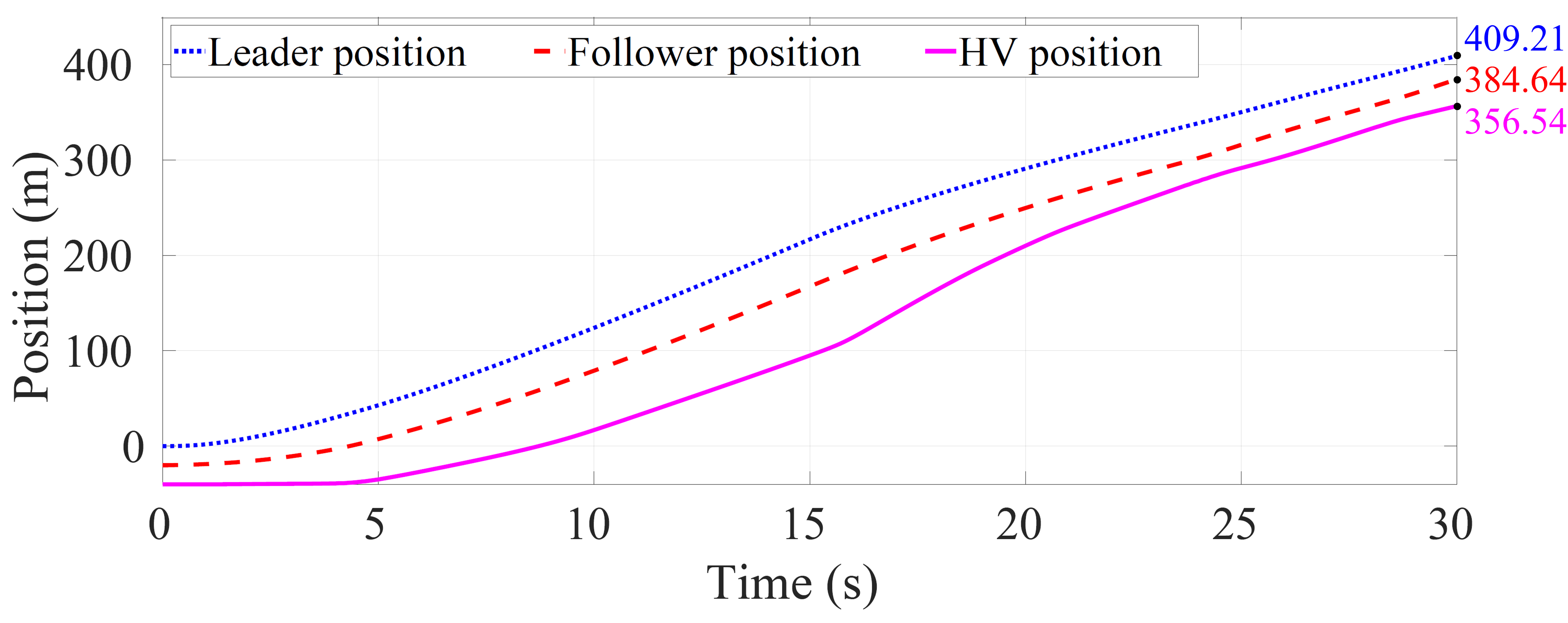} }}
    \qquad \qquad 
    \subfloat{{\includegraphics[trim=0.0cm 0cm 0.0cm 0.0cm, width=0.94\columnwidth]{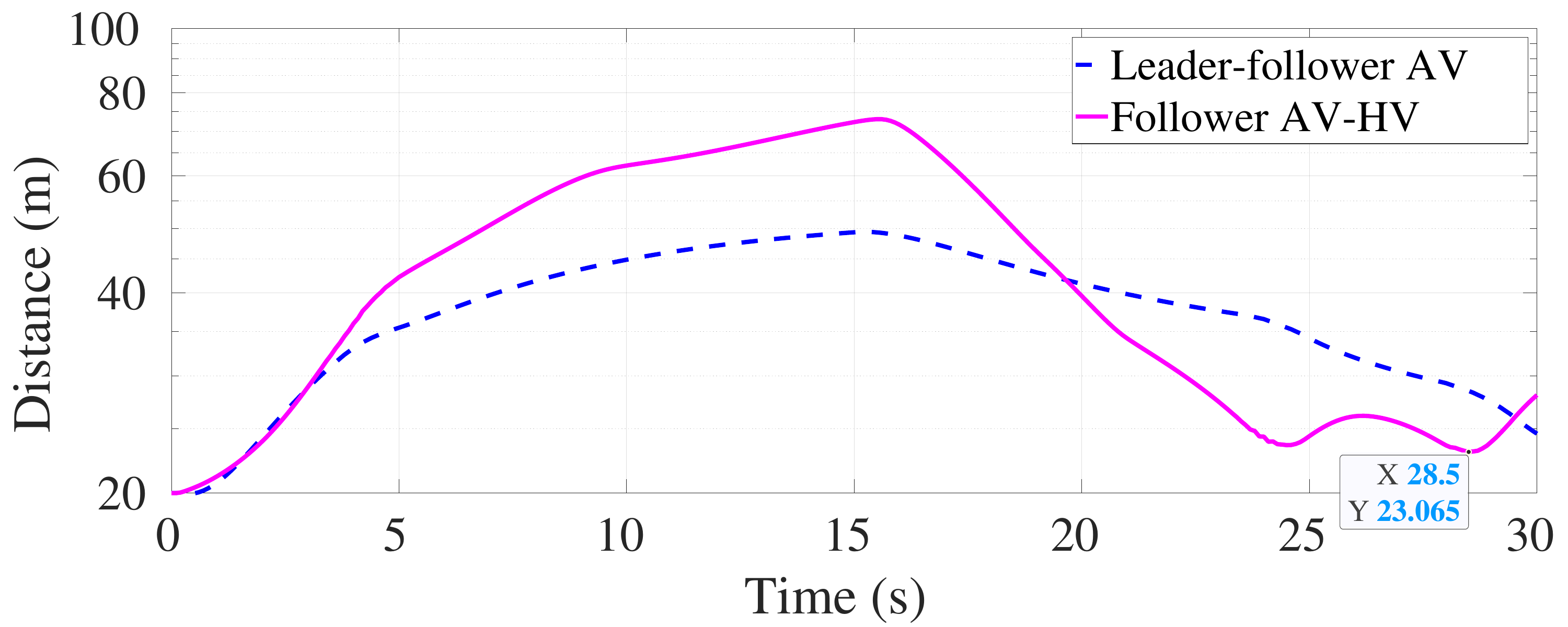} }}
    \caption{Results of an emergency braking simulation using the proposed GP-MPC.
    The graphs show, sequentially from top to bottom, the velocity response, position trajectories, and inter-vehicle distance. In comparison with the baseline nominal MPC, as shown in Fig. \ref{figure:nominal_simulation_braking}, GP-MPC enabled a two-meter increase in the minimum distance between the HV and the trailing AV. Moreover, as depicted in Tab. \ref{tab:simulation_metrics}, all vehicles covered larger distances under the GP-MPC model. Thus, the GP-MPC approach enhances safety without enforcing more stringent operational constraints than the nominal MPC, given that all vehicles travel a larger distance.} 
    \label{figure:gp_simulation_braking}
\end{figure}

\textbf{Emergency Braking Scenario:} 
In a sudden braking event, the leading AV's reference velocity $v^\text{ref}$ is initially defined as \(20 \, \text{m/s}\) and reduced to \(10 \, \text{m/s}\) at \(t=15\) s. The results are shown in Figs. \ref{figure:nominal_simulation_braking} and \ref{figure:gp_simulation_braking} for the nominal MPC and GP-MPC respectively, show a more pronounced difference in performance. 
Similar to the constant-speed tracking scenario, we plotted the velocity response, position trajectories, and inter-vehicle distance sequentially from top to bottom. The top velocity-response plots in Fig. \ref{figure:nominal_simulation_braking} and \ref{figure:gp_simulation_braking}, the deceleration of the leader and follower AVs is triggered at \(t=15\) s. As expected from autonomous control systems, the AVs decelerated immediately. In contrast, the HV exhibited a delayed response with notable acceleration for approximately 2 s before deceleration began at \(t=17\) s. This behavior was consistent with the HV model trained on the distracted driver data and reflects the delayed response times typical of human drivers, as discussed in Section \ref{sec:HV_model}. In real-world scenarios, even when drivers are attentive, their reaction times tend to be significantly longer than those of AVs, posing the risk of accidents and underscoring the necessity of accounting for the uncertainties linked to human-driven vehicles.

The difference in the overall velocity response between the nominal MPC and GP-MPC was too minor to be easily identified from the uppermost plots, whereas it could be observed in the middle plots. The middle plots show the position trajectories of all the vehicles. Under GP-MPC, all vehicles in the mixed-vehicle platoon achieved a greater traveled distance; thus, all vehicles moved at higher average speeds. The inter-vehicle distance plots (bottom graphs) demonstrate the capability of GP-MPC to maintain larger safety margins. At \(t=28.5\) s, GP-MPC ensured a minimum distance of \(23.065\) m between the vehicles, compared with \(20.684\) m under the nominal MPC. This increased distance is significant in scenarios in which the human reaction time is variable and uncertain. Plots of the velocity response, position trajectories, and inter-vehicle distance are shown in Figs. \ref{figure:nominal_simulation_braking} and \ref{figure:gp_simulation_braking} verify the superiority of GP-MPC in managing mixed-traffic platoons, particularly during emergency maneuvers where the unpredictability of human drivers is a significant factor. GP-MPC offers a reinforced safety net by preserving larger minimum distances between vehicles. Additionally, it facilitates a more fluid traffic flow by enabling higher travel speeds without compromising safety.

Table \ref{tab:simulation_metrics} consolidates these observations. Under constant velocity conditions (Tab. \ref{tab:simulation_metrics_con}), GP-MPC slightly increased the minimum distance between the second AV and the HV ($P_{fh}$) to 29.64 m, compared with 29.51 m for the nominal MPC. This modest improvement resulted from the integration of uncertainty assessments by GP-MPC, reflecting the relatively stable behavior of HVs in constant-velocity scenarios. However, in more dynamic scenarios, such as emergency braking (Tab. \ref{tab:simulation_metrics}), the effectiveness of GP-MPC is more pronounced. $P_{fh}$ for GP-MPC increased by approximately 2 m more than that of the nominal MPC. This significant difference highlights the effectiveness of GP-MPC in adapting to high-uncertainty scenarios, where its predictive capability enables a safer distance margin in more demanding traffic scenarios.
\begin{table}
    \caption{Results of emergency braking simulation: The location of each vehicle and the minimum relative distance of HV--AV. }
    \label{tab:simulation_metrics}
    \begin{center}
        \begin{tabular}{|c|c|c|c|c|}
            \hline Controller & AV 1 & AV 2 & HV & Min HV--AV  \\
            \hline Nominal MPC & 404.94 m & 373.08 m & 348.42 m & 20.68 m\\
            \hline GP-MPC & \textbf{409.21} m & \textbf{384.64} m &  \textbf{356.54} m & \textbf{23.07} m\\
            \hline
        \end{tabular}
    \end{center}
\end{table}

\subsection{Benchmark comparison}
We have demonstrated the enhanced accuracy of our GP+ARX model in predicting HV velocity compared to the nominal ARX model. Incorporating uncertainties in HVs quantified by GP models within our MPC framework has notably improved safety and efficiency in mixed-vehicle platooning. This section uses the constant speed model (CSM) as a benchmark to further highlight significant advancements of our method in modeling precision and safety guarantees within a mixed-traffic environment.

The choice of CSM for comparison is grounded in its compatibility with our velocity-based HV modeling method. Contrasting with the intelligent driver model (IDM) reviewed in Sec. \ref{sec:related_work},  and neural network models (NNM) like the one used in \cite{guo2021anticipative}, which depend on noisier acceleration data, both our method and CSM approaches are based on velocity without incorporating acceleration dynamics. This similarity ensures a more relevant and direct comparison. By defining the HV velocity equal to the AV ahead at every current time step $k$, a CSM model-based MPC strategy is developed as follows:
\begin{ceqn}
    \begin{align} 
       \text{Equation} \ \eqref{eqn:mpc_a} & \nonumber \\
       \text{with} \ \mathbb{V} = & \left\{v_{{i}|k}^1, v_{{i}|k}^\mathbf{n_a}, v_{{i}|k}^{H}, p_{{i}|k}^\mathbf{n_a}, p_{{i+1}|k}^{H}, \mathrm{acc}_{{i}|k}^{\mathbf{n_a}} \right\} \, \nonumber \\
        \text {subject to} \nonumber \\
        v^{H}_{{i}|k} &= {v}_{i | k}^{N_a} \, , \tag{22a} \label{eqn:mpc_cons_a} \\
        p_{{i+1}|k}^{H} &= p_{{i}|k}^{H} + T \, v_{{i}|k}^{N_a} \, , \tag{22b} \label{eqn:mpc_cons_b}\\
        p_{{i}|k}^{N_a} & - \mu_{{i}|k}^{p^{H}} \geq \Delta \, , \tag{22c} \label{eqn:mpc_cons_c}\\
        \text{Equa}&\text{tions} \ \eqref{eqn:mpc_b}, \eqref{eqn:mpc_c}, \eqref{eqn:mpc_g}, \text{and} \ \eqref{eqn:mpc_i}-\eqref{eqn:mpc_n} \nonumber \, .
    \end{align}
\end{ceqn}

We conducted constant-velocity tracking and emergency braking scenario simulations with the developed CSM model-based MPC strategy, applying the same settings detailed in Sec. \ref{sec:simulation_setup}. These results are detailed in Tab. \ref{tab:simulation_metrics_constVel} including the positional data and minimum relative distances, which revealed that while the CSM-based MPC can achieve stable platooning control, it displays limitations in complex scenarios like emergency braking. 

In the constant-velocity tracking scenarios, experiment results including the velocity response, position trajectories, and inter-vehicle distances are similar to those presented in Fig. \ref{figure:gp_simulation}. As shown in Tab. \ref{tab:simulation_metrics_constVel}, while the CSM-based MPC strategy demonstrated an ability to achieve steady-state control in platooning in this less demanding traffic condition, it notably led to shorter overall travel distances for all vehicles, indicating a generally lower average speed across the mixed platoon. 
\begin{table}
    \caption{CSM model-based MPC simulation results: The location of each vehicle and the minimum relative distance of the HV--AV platoon.
    }
    \label{tab:simulation_metrics_constVel}
    \begin{center}
        \begin{tabular}{|c|c|c|c|c|}
            \hline Simulations & AV 1 & AV 2 & HV & Min HV--AV  \\
            \hline Const. Vel. Track & 504.66 m & 442.19 m & 406.30 m & 35.89 m\\
            \hline Emergency Braking  & 394.24 m & 346.29 m & 349.55 m & -3.26 m\\
            \hline
        \end{tabular}
    \end{center}
\end{table}

\begin{figure}
    \centering
    \vspace{0.1cm}
    \subfloat{{\includegraphics[trim=0cm 0cm 0cm 0cm, width=0.94\columnwidth]{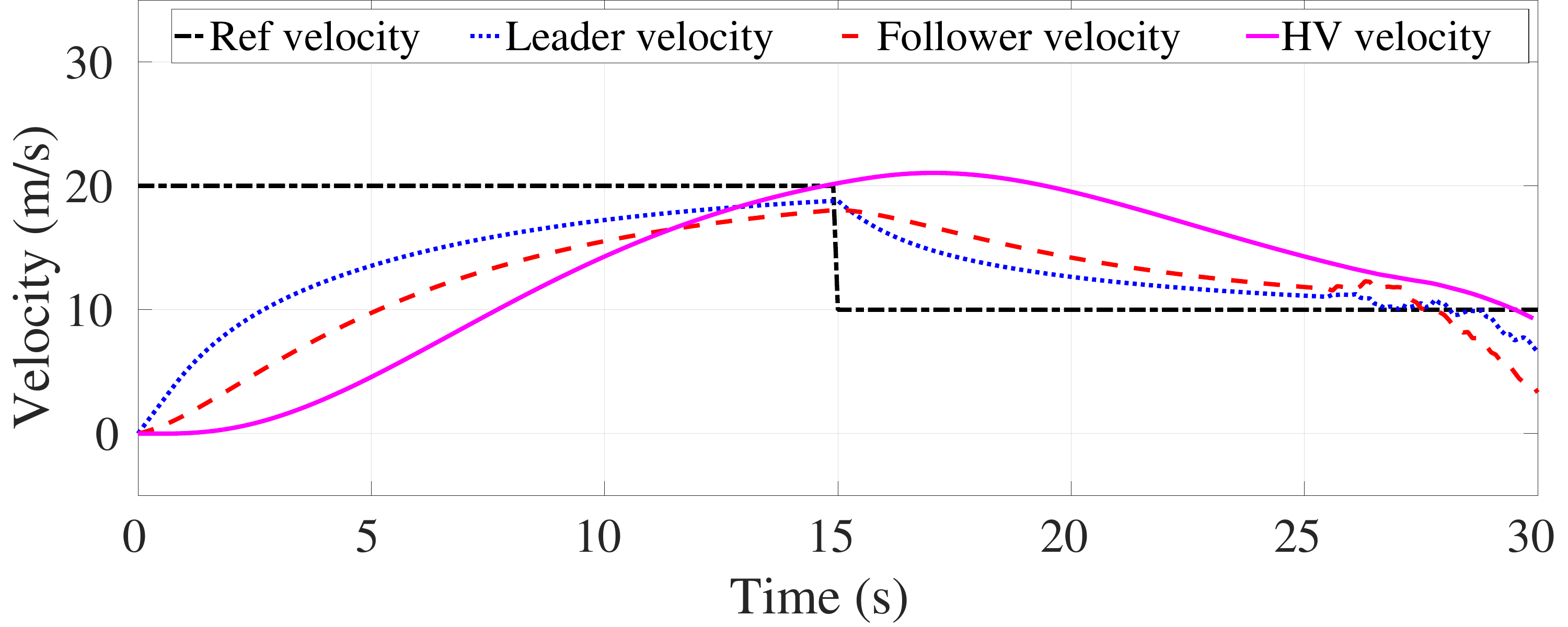} }}
    \qquad \qquad     
    \vspace{0.05cm}
    \subfloat{{\includegraphics[trim=0cm 0cm 0cm 0.0cm, width=0.94\columnwidth]{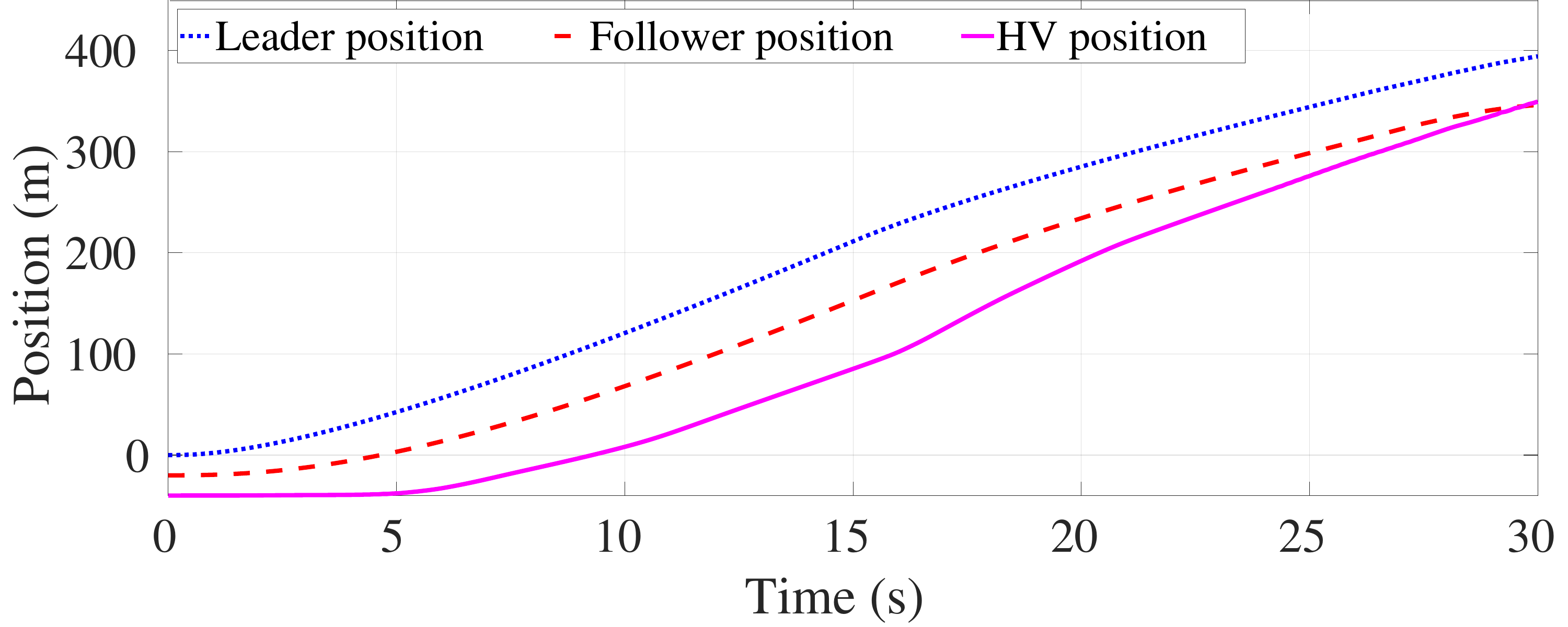} }}
    \qquad \qquad 
    \subfloat{{\includegraphics[trim=0.0cm 0cm 0.0cm 0.0cm, width=0.94\columnwidth]{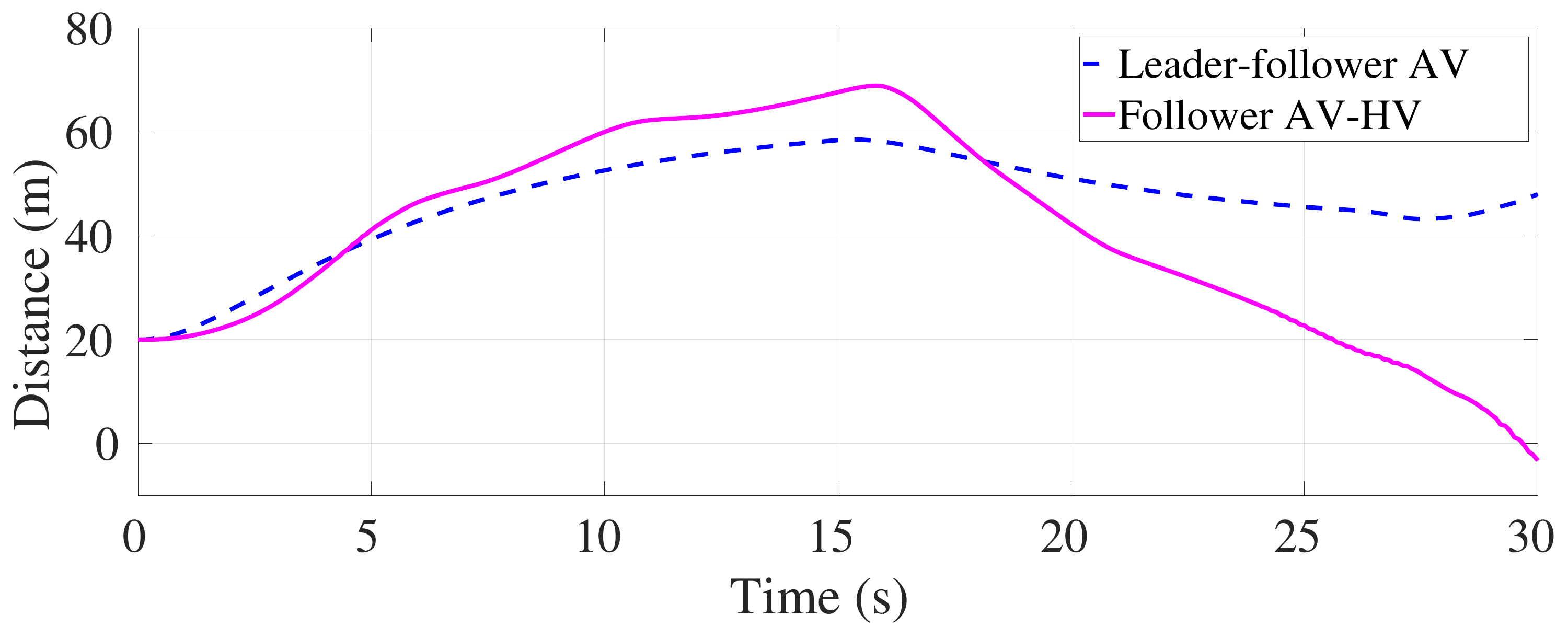} }}
    \caption{Results of an emergency braking simulation using the CSM based-MPC. The graphs show, sequentially from top to bottom, the velocity response, position trajectories, and inter-vehicle distance. The rear-end collision of the HV with the preceding AV underscores the CSM's inadequacy in accurately predicting HV behavior, thereby challenging its ability to ensure safe control in dynamic mixed-traffic situations.} 
    \label{figure:gp_simulation_braking_constVel}
\end{figure}

Fig. \ref{figure:gp_simulation_braking_constVel} illustrates the limitations of the CSM model-based MPC strategy in managing emergency braking scenarios. The plots of position trajectories and inter-vehicle distances reveal a critical issue: the HV fails to prevent a rear collision with the AV ahead. This outcome underscores the CSM's deficiency in accurately modeling HV behavior in high-stakes situations. In such complex traffic scenarios, the model's inability to effectively predict and react to rapid changes proves to be a significant drawback, highlighting the need for more sophisticated modeling techniques to ensure safe and effective traffic control. \\

\subsection{Summary and discussions}
\label{sec:summary_discussions}
This section presents a thorough evaluation of our GP-MPC strategy through detailed simulations in mixed-traffic settings and compares its performance with a baseline MPC. We focused on scenarios that included constant-velocity tracking and emergency braking to demonstrate the efficacy and practical applicability of the model.

In constant-velocity tracking, GP-MPC exhibited precision in aligning AVs with the reference speed and effectively influenced HV behavior towards a desirable traffic flow despite initial disturbances. This demonstrated the adaptability and resilience of the model in dealing with unpredictable HV behaviors. Emergency braking simulations further highlighted the strengths of GP-MPC. It maintains larger safety margins between vehicles, which is a critical aspect in emergencies involving variable human driver reactions. This capability to accommodate uncertainties while ensuring safe distances emphasizes the robustness of GP-MPC as a control strategy in mixed-traffic environments. Moreover, GP-MPC was effective in enhancing traffic flow efficiency. In both scenarios, vehicles under GP-MPC achieved greater traveled distances compared with the nominal MPC, indicating higher average speeds without compromising safety. This balance between safety and efficiency makes GP-MPC a promising approach for real-world traffic management.

Our benchmark analysis also compellingly validates the superiority of our novel GP+ARX model over traditional HV modeling methods. By effectively integrating GP models to quantify uncertainties within the MPC framework, our approach not only enhances accuracy in velocity prediction but also significantly improves safety and efficiency in mixed-vehicle platooning. The comparison with the CSM-based MPC underscores our method's advanced capability in handling complex traffic scenarios, thereby confirming the substantial improvements of our proposed technique. 

However, a challenge encountered was the direct quantitative comparison between our GP-MPC and other advanced MPC strategies for mixed-traffic applications, as discussed in Sec. \ref{sec:related_work}. These strategies, each with its unique objectives and assumptions, consider different aspects of mixed-traffic control. This diversity necessitates a nuanced understanding of the strengths and limitations of each method and underscores the need for tailored solutions in mixed-traffic management. The integration of GP-MPC with other advanced MPC strategies offers exciting possibilities. Such collaborations could merge the robustness of the probabilistic bounds in Tube MPC \cite{feng2021robust} and the scalability of distributed MPC \cite{zhan2022data} with the capabilities of our GP-MPC in uncertainty management. This synthesis can result in more holistic and adaptive control strategies, thereby significantly enhancing the safety and efficiency of mixed-traffic systems. This collaborative effort can enable resilient and responsive traffic management solutions, marking significant advancements in the field.

\section{Conclusion}
\label{sec:conclusion}

In this study, we addressed the challenge of safely controlling a mixed-vehicle platoon comprising both autonomous vehicles (AVs) and human-driven vehicles (HVs) in longitudinal car-following scenarios. We introduced a hybrid modeling approach that combines a first-principles model with a Gaussian process (GP) learning-based model. This innovative integration not only enhances the accuracy of predicting HV behavior but also quantifies the uncertainty associated with HVs in car-following scenarios. Leveraging this model, we developed a model predictive control (MPC) strategy focused on enhancing safety within mixed-vehicle platoons. Simulation studies comparing our method with a baseline MPC demonstrated that our approach not only provides stronger safety assurance but also facilitates more efficient motion behaviors for all vehicles in the platoon.

Our study had certain limitations and areas for future research. One limitation is the focus on longitudinal car-following scenarios, specifically with HVs following an AV platoon. Extending this research to encompass more complex traffic conditions, such as HVs positioned within an AV platoon or executing merging and lane-changing maneuvers, can provide broader insights. Another potential area for enhancement is the diversity of the dataset. Currently, the dataset for the hybrid HV model exhibits limited variation. Collecting more diverse data from a wider range of drivers and driving conditions can further improve the accuracy of the model. Although the GP model has yielded promising results, it is currently trained offline. Developing more computation-efficient methodologies such as sparse GP or local GP models can make our solution more applicable to real-world vehicle applications. Lastly, integrating our GP-MPC with other advanced traffic management methods could further enhance the control efficiency and adaptability for mixed traffic. This collaborative approach can result in more responsive and comprehensive traffic solutions, laying the foundation for advanced traffic management methodologies.

\section{Acknowledgements}

This project received partial support from Magna International and the Discovery Grant from the Natural Sciences and Engineering Research Council of Canada.

\bibliographystyle{elsarticle-num} 
\bibliography{bibliography}

\begin{thebibliography}{10}
\expandafter\ifx\csname url\endcsname\relax
  \def\url#1{\texttt{#1}}\fi
\expandafter\ifx\csname urlprefix\endcsname\relax\def\urlprefix{URL }\fi
\expandafter\ifx\csname href\endcsname\relax
  \def\href#1#2{#2} \def\path#1{#1}\fi

\bibitem{guanetti2018}
J.~Guanetti, Y.~Kim, F.~Borrelli, Control of connected and automated vehicles: State of the art and future challenges, Annual reviews in control 45 (2018) 18--40.

\bibitem{martinez2021}
M.~Mart{\'\i}nez-D{\'\i}az, C.~Al-Haddad, F.~Soriguera, C.~Antoniou, Platooning of connected automated vehicles on freeways: a bird’s eye view, Transportation research procedia 58 (2021) 479--486.

\bibitem{aramrattana2021}
M.~Aramrattana, A.~Habibovic, C.~Englund, Safety and experience of other drivers while interacting with automated vehicle platoons, Transportation research interdisciplinary perspectives 10 (2021) 100381.

\bibitem{rahmati2019}
Y.~Rahmati, M.~Khajeh~Hosseini, A.~Talebpour, B.~Swain, C.~Nelson, Influence of autonomous vehicles on car-following behavior of human drivers, Transportation research record 2673~(12) (2019) 367--379.

\bibitem{petrovic2020}
P.~Dorde, M.~Radomir, D.~Pesic, Traffic accidents with autonomous vehicles: type of collisions, manoeuvres and errors of conventional vehicles’ drivers, Transportation research procedia 45 (2020) 161--168.

\bibitem{sadat2020}
A.~Sadat, S.~Casas, M.~Ren, X.~Wu, P.~Dhawan, R.~Urtasun, Perceive, predict, and plan: Safe motion planning through interpretable semantic representations, in: European Conference on Computer Vision, Springer, 2020, pp. 414--430.

\bibitem{macadam2003}
C.~C. Macadam, {Understanding and Modeling the Human Driver}, Vehicle System Dynamics (2003).

\bibitem{yu2022researches}
L.~Yu, R.~Wang, Researches on adaptive cruise control system: A state of the art review, Proceedings of the Institution of Mechanical Engineers, Part D: Journal of Automobile Engineering 236~(2-3) (2022) 211--240.

\bibitem{li2022cooperative}
K.~Li, J.~Wang, Y.~Zheng, Cooperative formation of autonomous vehicles in mixed traffic flow: Beyond platooning, IEEE Transactions on Intelligent Transportation Systems 23~(9) (2022) 15951--15966.

\bibitem{yu2021}
S.~Yu, M.~Hirche, Y.~Huang, H.~Chen, F.~Allg{\"o}wer, Model predictive control for autonomous ground vehicles: A review, Autonomous Intelligent Systems 1~(1) (2021) 1--17.

\bibitem{wang2023learning}
J.~Wang, M.~T. Fader, J.~A. Marshall, Learning-based model predictive control for improved mobile robot path following using {Gaussian processes} and feedback linearization, Journal of Field Robotics 40 (2023) 1014--1033.

\bibitem{guo2021anticipative}
L.~Guo, Y.~Jia, Anticipative and predictive control of automated vehicles in communication-constrained connected mixed traffic, IEEE Transactions on Intelligent Transportation Systems 23~(7) (2021) 7206--7219.

\bibitem{li2023survey}
J.~Li, C.~Yu, Z.~Shen, Z.~Su, W.~Ma, A survey on urban traffic control under mixed traffic environment with connected automated vehicles, Transportation research part C: emerging technologies 154 (2023) 104258.

\bibitem{guo2020}
L.~Guo, Y.~Jia, Inverse model predictive control {(IMPC)} based modeling and prediction of human-driven vehicles in mixed traffic, IEEE Transactions on Intelligent Vehicles 6~(3) (2020) 501--512.

\bibitem{panwai2007}
S.~Panwai, H.~Dia, Neural agent car-following models, IEEE Transactions on Intelligent Transportation Systems 8~(1) (2007) 60--70.

\bibitem{khodayari2012}
A.~Khodayari, A.~Ghaffari, R.~Kazemi, R.~Braunstingl, A modified car-following model based on a neural network model of the human driver effects, IEEE Transactions on Systems, Man, and Cybernetics-Part A: Systems and Humans 42~(6) (2012) 1440--1449.

\bibitem{morton2016}
J.~Morton, T.~A. Wheeler, M.~J. Kochenderfer, Analysis of recurrent neural networks for probabilistic modeling of driver behavior, IEEE Transactions on Intelligent Transportation Systems 18~(5) (2016) 1289--1298.

\bibitem{lefevre2014a}
S.~Lefevre, Y.~Gao, D.~Vasquez, H.~E. Tseng, R.~Bajcsy, F.~Borrelli, Lane keeping assistance with learning-based driver model and model predictive control, in: 12th International Symposium on Advanced Vehicle Control, 2014.

\bibitem{qu2017}
T.~Qu, S.~Yu, Z.~Shi, H.~Chen, Modeling driver's car-following behavior based on hidden markov model and model predictive control: A cyber-physical system approach, in: 2017 11th Asian Control Conference (ASCC), IEEE, 2017, pp. 114--119.

\bibitem{lefevre2014b}
S.~Lef{\`e}vre, C.~Sun, R.~Bajcsy, C.~Laugier, Comparison of parametric and non-parametric approaches for vehicle speed prediction, in: 2014 American Control Conference, IEEE, 2014, pp. 3494--3499.

\bibitem{hewing2020}
L.~Hewing, K.~P. Wabersich, M.~Menner, M.~N. Zeilinger, Learning-based model predictive control: Toward safe learning in control, Annual Review of Control, Robotics, and Autonomous Systems 3 (2020) 269--296.

\bibitem{chen2010}
X.~Chen, L.~Li, Y.~Zhang, A {Markov} model for headway/spacing distribution of road traffic, IEEE Transactions on Intelligent Transportation Systems 11~(4) (2010) 773--785.

\bibitem{feng2021robust}
S.~Feng, Z.~Song, Z.~Li, Y.~Zhang, L.~Li, Robust platoon control in mixed traffic flow based on tube model predictive control, IEEE Transactions on Intelligent Vehicles 6~(4) (2021) 711--722.

\bibitem{zhan2022data}
J.~Zhan, Z.~Ma, L.~Zhang, Data-driven modeling and distributed predictive control of mixed vehicle platoons, IEEE Transactions on Intelligent Vehicles 8~(1) (2022) 572--582.

\bibitem{pirani2022}
M.~Pirani, Y.~She, R.~Tang, Z.~Jiang, Y.~V. Pant, Stable interaction of autonomous vehicle platoons with human-driven vehicles, in: 2022 American Control Conference (ACC), IEEE, 2022, pp. 633--640.

\bibitem{brezinski1994pade}
C.~Brezinski, J.~Van~Iseghem, Pad{\'e} approximations, Handbook of Numerical Analysis 3 (1994) 47--222.

\bibitem{sekizawa2007modeling}
S.~Sekizawa, S.~Inagaki, T.~Suzuki, S.~Hayakawa, N.~Tsuchida, T.~Tsuda, H.~Fujinami, Modeling and recognition of driving behavior based on stochastic switched {ARX} model, IEEE Transactions on Intelligent Transportation Systems 8~(4) (2007) 593--606.

\bibitem{wang2023intuitive}
J.~Wang, An intuitive tutorial to {Gaussian} process regression, Computing in Science \& Engineering 25~(4) (2023) 4--11.

\bibitem{brunke2021}
L.~Brunke, M.~Greeff, A.~W. Hall, Z.~Yuan, S.~Zhou, J.~Panerati, A.~P. Schoellig, Safe learning in robotics: From learning-based control to safe reinforcement learning, Annual Review of Control, Robotics, and Autonomous Systems 5 (2021).

\bibitem{rasmussen2006}
C.~E. Rasmussen, C.~K.~I. Williams, {Gaussian} processes in machine learning, MIT Press, 2006.
\newblock \href {https://doi.org/10.1142/S0129065704001899} {\path{doi:10.1142/S0129065704001899}}.

\bibitem{zhao2023safety}
C.~Zhao, H.~Yu, T.~G. Molnar, Safety-critical traffic control by connected automated vehicles, Transportation research part C: emerging technologies 154 (2023) 104230.

\bibitem{hewing2019}
L.~Hewing, J.~Kabzan, M.~N. Zeilinger, Cautious model predictive control using {Gaussian} process regression, IEEE Transactions on Control Systems Technology 28~(6) (2019) 2736--2743.

\end{thebibliography}

\end{document}